\pdfoutput=1

\documentclass[11pt]{article}

\usepackage[preprint]{acl}

\usepackage{times}
\usepackage{latexsym}

\usepackage[T1]{fontenc}

\usepackage[utf8]{inputenc}

\usepackage{microtype}

\usepackage{inconsolata}

\usepackage{longtable} 

\usepackage[utf8]{inputenc} 
\usepackage[T1]{fontenc}    
\usepackage{hyperref}       
\usepackage{url}            
\usepackage{booktabs}       
\usepackage{amsfonts}       
\usepackage{nicefrac}       
\usepackage{microtype}      
\usepackage{graphicx} 
\usepackage{diagbox}       
\usepackage{float} 
\usepackage{wrapfig}  
\bibstyle{acl_natbib}
\usepackage{colortbl}     
\usepackage{graphicx}     
\usepackage{amsmath}      
\usepackage{colortbl}
\usepackage{tcolorbox}
\usepackage{multirow}
\usepackage{makecell}
\usepackage{bbding}
\usepackage{longtable}  

\newcommand{\cellcolorheatblue}[1]{
    \ifdim #1 pt > 50 pt
        \cellcolor{blue!40!white}#1
    \else\ifdim #1 pt > 45 pt
        \cellcolor{blue!35!white}#1
    \else\ifdim #1 pt > 40 pt
        \cellcolor{blue!30!white}#1
    \else\ifdim #1 pt > 35 pt
        \cellcolor{blue!25!white}#1
    \else\ifdim #1 pt > 30 pt
        \cellcolor{blue!20!white}#1
    \else\ifdim #1 pt > 25 pt
        \cellcolor{blue!15!white}#1
    \else\ifdim #1 pt > 20 pt
        \cellcolor{blue!10!white}#1
    \else\ifdim #1 pt > 15 pt
        \cellcolor{blue!5!white}#1
    \else\ifdim #1 pt > 10 pt
        \cellcolor{blue!0!white}#1
    \else\ifdim #1 pt > 5 pt
        \cellcolor{blue!0!white}#1
    \else
        \cellcolor{white}#1
    \fi\fi\fi\fi\fi\fi\fi\fi\fi\fi
}
\newcommand{\cellcolorheatblueblue}[1]{
    \ifdim #1 pt > 17 pt
        \cellcolor{blue!40!white}#1
    \else\ifdim #1 pt > 15 pt
        \cellcolor{blue!35!white}#1
    \else\ifdim #1 pt > 13 pt
        \cellcolor{blue!30!white}#1
    \else\ifdim #1 pt > 11 pt
        \cellcolor{blue!25!white}#1
    \else\ifdim #1 pt > 9 pt
        \cellcolor{blue!20!white}#1
    \else\ifdim #1 pt > 7 pt
        \cellcolor{blue!15!white}#1
    \else\ifdim #1 pt > 5 pt
        \cellcolor{blue!10!white}#1
    \else\ifdim #1 pt > 3 pt
        \cellcolor{blue!5!white}#1
    \else\ifdim #1 pt > 2 pt
        \cellcolor{blue!0!white}#1
    \else\ifdim #1 pt > 1 pt
        \cellcolor{blue!0!white}#1
    \else
        \cellcolor{white}#1
    \fi\fi\fi\fi\fi\fi\fi\fi\fi\fi
}
\usepackage{adjustbox} 

\usepackage{listings}

\newcommand{\cellcolorheat}[1]{
    \ifdim #1 pt >95 pt
        \cellcolor{red!90!white}#1
    \else\ifdim #1 pt > 90 pt
        \cellcolor{red!80!white}#1
    \else\ifdim#1 pt > 85 pt
        \cellcolor{red!70!white}#1
    \else\ifdim #1 pt > 80 pt
        \cellcolor{red!60!white}#1
    \else\ifdim #1 pt > 75 pt
        \cellcolor{red!55!white}#1
    \else\ifdim#1 pt > 70 pt
        \cellcolor{red!50!white}#1     
    \else\ifdim #1 pt > 65 pt
        \cellcolor{red!45!white}#1
    \else\ifdim #1 pt > 60 pt
        \cellcolor{red!40!white}#1
    \else\ifdim #1 pt > 55 pt
        \cellcolor{red!35!white}#1
    \else\ifdim #1 pt > 50 pt
        \cellcolor{red!30!white}#1    
    \else\ifdim #1 pt > 45 pt
        \cellcolor{red!25!white}#1
    \else\ifdim #1 pt > 40 pt
        \cellcolor{red!20!white}#1
    \else\ifdim #1 pt > 35 pt
        \cellcolor{red!15!white}#1
    \else\ifdim #1 pt > 30 pt
        \cellcolor{red!10!white}#1
    \else\ifdim #1 pt > 25 pt
        \cellcolor{red!6!white}#1
    \else\ifdim #1 pt > 20 pt
        \cellcolor{red!4!white}#1
    \else\ifdim #1 pt > 15 pt
        \cellcolor{red!2!white}#1
     \else\ifdim #1 pt > 10 pt
        \cellcolor{red!1!white}#1
    \else
        \cellcolor{white}#1
    \fi\fi\fi\fi\fi\fi\fi\fi\fi\fi\fi\fi\fi\fi\fi\fi\fi
}

\title{$C^3$-Bench: The Things Real Disturbing LLM based Agent in Multi-Tasking}

\author{
 \textbf{Peijie Yu}{\textsuperscript{1}\textsuperscript{*}\textsuperscript{\textdagger}},
 \textbf{Yifan Yang}{\textsuperscript{1}\textsuperscript{*}\textsuperscript{\textdagger}},
 \textbf{Jinjian Li}{\textsuperscript{1}\textsuperscript{*}},
 \textbf{Zelong Zhang\textsuperscript{1}},
\\
 \textbf{Haorui Wang\textsuperscript{1}},
 \textbf{Xiao Feng\textsuperscript{1}},
  \textbf{Feng Zhang\textsuperscript{1}}
\\
\\
 \textsuperscript{1}Tencent HunYuan Team
\\
 \small{
   \textbf{Correspondence:} \href{mailto:email@domain}{\{peijieyu, ioanyang\}@tencent.com}
 }
}

\begin{document}
\maketitle

{
\renewcommand{\thefootnote}{*}
\footnotetext[1]{Equal Contributions.}
\renewcommand{\thefootnote}{\textdagger}
\footnotetext[2]{Corresponding authors.}
}

\begin{abstract}
Agents based on large language models leverage tools to modify environments, revolutionizing how AI interacts with the physical world. Unlike traditional NLP tasks that rely solely on historical dialogue for responses, these agents must consider more complex factors, such as inter-tool relationships, environmental feedback and previous decisions, when making choices. Current research typically evaluates agents via multi-turn dialogues. However, it overlooks the influence of these critical factors on agent behavior. To bridge this gap, we present an open-source and high-quality benchmark $C^3$-Bench. This benchmark integrates attack concepts and applies univariate analysis to pinpoint key elements affecting agent robustness. In concrete, we design three challenges: navigate complex tool relationships, handle critical hidden information and manage dynamic decision paths. Complementing these challenges, we introduce fine-grained metrics, innovative data collection algorithms and reproducible evaluation methods. Extensive experiments are conducted on 49 mainstream agents, encompassing general fast-thinking, slow-thinking and domain-specific models. We observe that agents have significant shortcomings in handling tool dependencies, long context information dependencies and frequent policy-type switching. In essence, $C^3$-Bench aims to expose model vulnerabilities through these challenges and drive research into the interpretability of agent performance. The benchmark is publicly available at https://github.com/Tencent-Hunyuan/C3-Benchmark.
\end{abstract}

\section{Introduction}

Large language models (LLMs) are evolving rapidly and agents built on them have become a research hotspot\cite{Gemini,guo2025deepseek,zeng2025toolace}. These agents interact with the real world through various tools, opening up new avenues for AI development.

A typical agent task execution process \cite{yao2023react} involves five factors. Users propose tasks (User) with relevant tools (Tool). Based on them, the agent reasonably takes actions (Action) and observes the feedback of the actions (Observation). One task often involves multiple Action-Observation steps. And task completion with a feedback to the user (Summary). The user then triggers a new task cycle and this becomes a multi-tasking scenario. Sequential task execution poses more complex decision making challenges for agents compared to traditional language tasks, involving new factors such as tool dependencies, environmental feedback and historical decisions. However, existing research \cite{huang2024planning,qintoolllm,duanytool,yao2024tau,gorilla-openfunctions-v3} fails to fully recognize the differences between agent multi-tasking and traditional multi-turn dialogues and rarely explores how these factors impact agent decision making.

To address this, we propose an open-source multi-task agent benchmark $C^3$-Bench. It introduces attack concepts (non-security-focused) to reveal model weaknesses, uses univariate analysis to study factors affecting agent robustness, and challenges agents by increasing complex factor types and frequencies. For the five roles in multi-tasking (user, tool, action, observation, summary), three challenges and fine-grained evaluation metrics are designed.

\begin{figure*}[!t]
  \centering
  \includegraphics[width=1.0\linewidth]{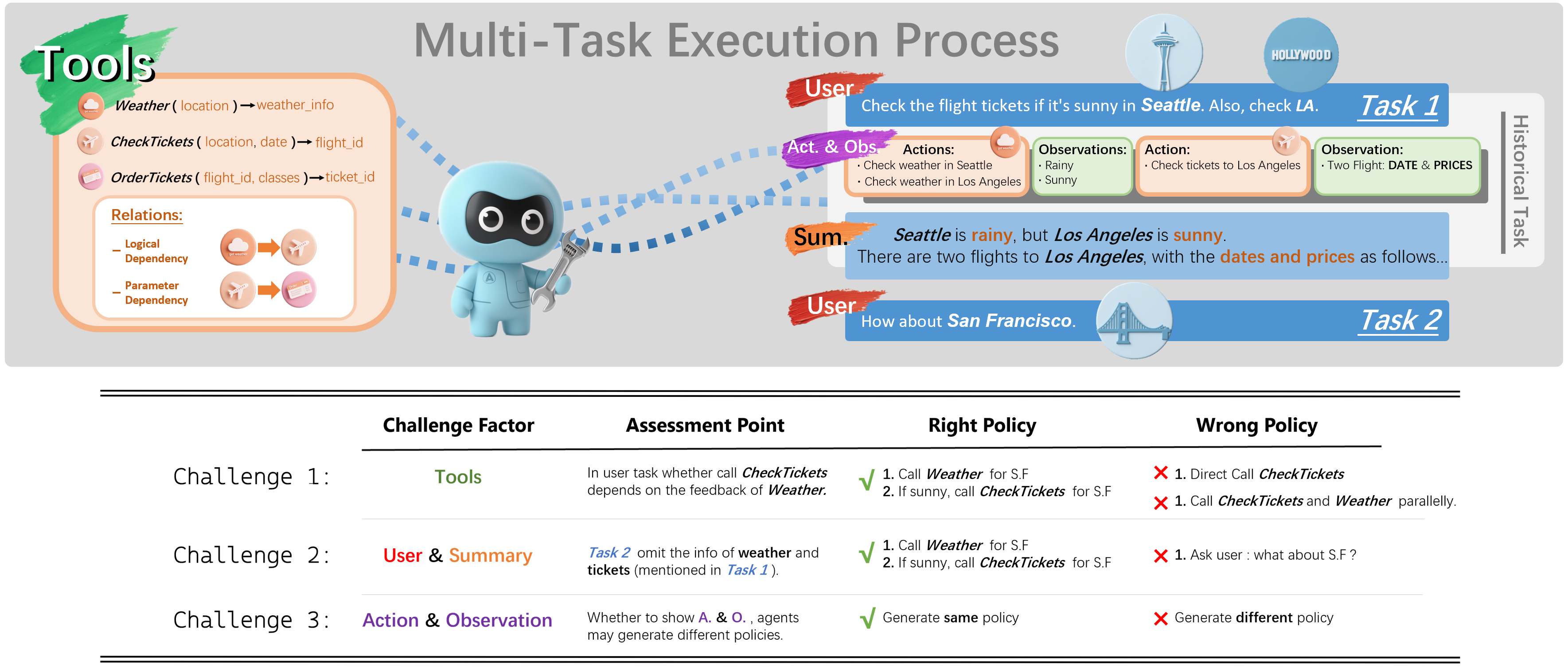} 
  \caption {
User wants to make a travel plan with good weather and suitable flights. He initially had \textbf{two} candidate cities and later \textbf{added one}. $C^3$-Bench poses three challenges for Agents based on the \textbf{five factors} of the execution proceed. }

  \label{fig:exp_1}

\end{figure*}

\textbf{Challenge 1 on \textcolor{teal}{Tools}:}
One key sign of human intelligence is the ability to understand and use different tools; thus, we first study the agent's capabilities in this regard. We choose multi-task data that need multiple tools, and test agent accuracy in two situations: with tool dependencies and without them. Our evaluation assesses not only the correctness of tool-call policies but also the accomplish progress and efficiency.

\textbf{Challenge 2 on \textcolor{red}{User} \& \textcolor{orange}{Summary}:}
After verifying how tools affect decision making, we study the impact of related tasks. First, we focus on shared information between tasks. All tasks in this challenge require agents to obtain information from previous conversations. We create three strategies to hide information and evaluate their affect on model accuracy. We test agents using only the user task and summary parts of past dialogues. And each data set has 2-4 related tasks, so we also investigate the influence of task number on model accuracy.

\textbf{Challenge 3 on \textcolor{violet}{Action} \& \textcolor{violet}{Observation}
:} Just as a rugby coach's playbook, agents need diverse and flexible policies to complete user tasks. However, variable historical strategies affect current decision making. This time, we focus on the impact of historical decisions. Based on Challenge 2, we add task execution details to the old dialogues to induce the agent to generate a different policy. To test the impact of execution processes on models, we first expand the agent's policy space (4 policy types). Then, we gather tasks to traverse all policy changing paradigms across 2 to 4 tasks (336 types in total). Finally, we compare the results of Challenge 2 and 3 to analyze the impact of decision processes on agents. We find that larger result changes correlate with poorer model performance. Therefore, we further propose new metrics to pinpoint the factors that affect model stability. 

We analyze abundant real user logs \cite{sharegpt} to design the benchmark, yet all testing data do not come from user logs. Five human experts collaborate with an innovative multi-role data framework to produce high-quality test data. The leaderboard, test framework and data generation framework are open-source.

Extensive experiments on 49 mainstream SOTA models shows that the best performing model scores only \textbf{55.34}, highlighting the limitations of LLM-based agents. Analysis indicates that the three factors emphasized by the benchmark lead to a decrease in model performance. With proposed fine-grained metrics, we identify the most impactful issues on the model: inter-tool dependencies (causes \textbf{28.7\%} accuracy declines), long-context information dependencies (causes \textbf{11.0\%} accuracy declines), the number of tasks (one more task causes \textbf{10.1\%} accuracy declines) and the frequency of policy switching in historical decisions  (influences \textbf{68\% }models). These insights expose the inconsistent development of model capabilities, offering potential directions for improving model interpretability and improving model performance.

\section{$C^3$-Bench}

\subsection{Formulation}
\textbf{Process Modeling} We model LLM-based agents' task processing as a specialized Markov decision process, denoted $<g, A, O, s, \theta>$, where $g$ is the user task, $A$ the action space, $O$ the observation space, $s$ the agent's summary and $\theta$ the agent's parameters.

\textbf{Single-task Decision} For a single task, the agent generates policies from the user task $g$, iteratively calling tools and observing the environment. The $t$-step \textbf{policy} is $P = \{a_1, o_1, ..., o_t\}$ and the task summary is $s=\theta(g, P)$ after completion.

\textbf{Multi-task Decision} In multi-task scenarios, the agent integrates current and historical task information for decision making: $P_i=\theta(g_i | {g_1, P_1, s_1, ..., s_{i-1}})$. In practical applications, the environment often provides abundant information, such as  data from multiple web pages. To enable fast responses and minimize input, the agent always conceals task execution history, showing only the requirements and results. Therefore, in challenge 2 the historical decisions are omitted: $P_i=\theta(g_i | {g_1, s_1, ..., s_{i-1}})$.

\textbf{Policy Categorization} Based on real-world dialogues, we classify agent policies into four types: single-tool calling ($P_{single}$), multi-tool calling ($P_{multi}$), chatting with the user ($P_{chat}$) and tool calling after parameter clarification ($P_{clarify}$). To cover all application cases, we divide $P_{multi}$ into more fine-grained categories to examine the model: serial ($P_{multi}^{S}$), parallel ($P_{multi}^{P}$) and combined ($P_{multi}^{S + P}$) execution. $P_{multi}^{S} $ and $P_{multi}^{S + P}$ indicate that tools have dependencies. We show demos in Appendix \ref{demo-1}, \ref{demo-2}.

\textbf{Policy Transition Frequency} We define Policy Transition Frequency ($PTF$) to measure the complexity of decision trajectory. For a task sequence $\mathbb{G}_n = \{g_1, g_2, ..., g_n\}$ and its policy sequence $\mathbb{P}_n= \{P_1, P_2, ..., P_n\}$, $PTF$ is formulated as:
\begin{equation}
PTF(\mathbb{P}_n) = \sum_{i=1}^{n-1} \mathbb{I}(Type(P_i) \neq Type(P_{i+1}))
\end{equation}
where $\mathbb{I}(\cdot)$ is the indicator function, returns 1 when the condition is met and 0 otherwise. Higher $PTF$ values indicate more frequent policy changes, and do not directly relate to the number of tasks. For example, $\{P_{single},P_{single},P_{single}\}$ has $PTF = 0$, while $\{P_{single},P_{multi},P_{single}\}$ has $PTF = 2$. We show a demo in Appendix \ref{demo-4}.

\subsection{Challenge 1: Complex Tool Relationships}
External tools are the core feature that differentiates agent tasks from others. We first test models' ability to understand and use tools. With enhanced planning capabilities, users increase the expectations of agents to combine tools to complete complex tasks. Therefore. we select tasks related to agent policy $P_{multi}$ from the proposed benchmark as test data, each entry contains 1-4 tasks. As mentioned above, these tasks are divided into three subcategories: $P_{multi}^{S}$, $P_{multi}^{P}$ and $P_{multi}^{S + P}$, mainly due to parameter or logical dependencies among tools. A capable agent should recognize these dependencies during decision making, plan the correct sequence of tool calls and combine independent steps.

\begin{wrapfigure}{r}{0.21\textwidth} 
  \centering
  \includegraphics[width=\linewidth]{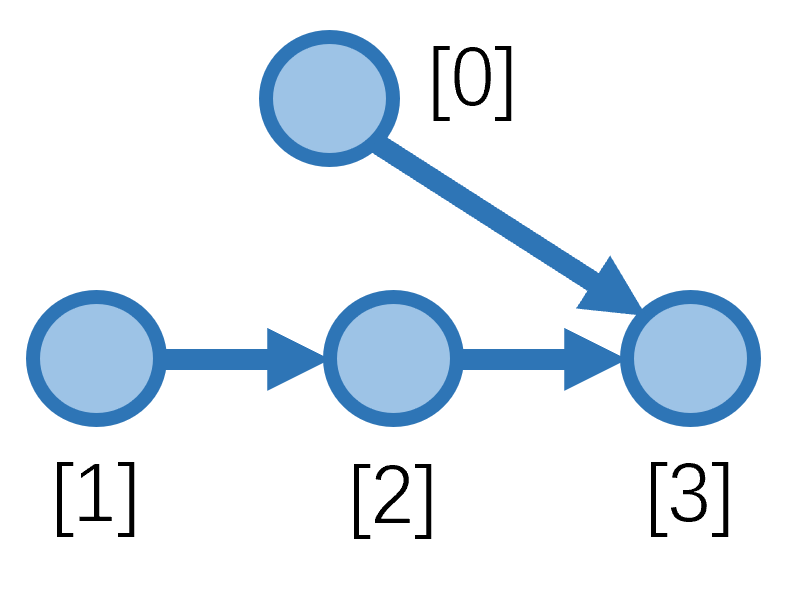} 
  \caption {Relationships among various tools.}
  \label{fig:toy_1}
\end{wrapfigure}
\textbf{Evaluation Method} 

The $P_{multi}$ tasks may have multiple correct decision making trajectories. We first propose a novel evaluation method to assess the correctness of the trajectories. Then we introduce fine-grained metrics to assess the agent's accomplish progress in $P_{multi}^{S}, P_{multi}^{S + P}$ tasks and policy's efficacy in $P_{multi}^{P}, P_{multi}^{S + P}$ tasks.

\textbf{Correctness}
First, we manually determine the tool dependencies. Then, we apply a depth-first search-based topological sorting algorithm to generate all possible execution paths. Our approach outperforms previous methods \cite{qiao2024benchmarking, shen2024taskbench} by constructing complete optimal and suboptimal sequences, rather than limited suboptimal ones.

During agent testing, we use an incremental path matching strategy in a decision tree. Each agent action either terminates the path if mismatched or prunes the sub-tree for valid actions, reducing future options.

\begin{figure*}[h]
\centering
\includegraphics[width=0.8\linewidth]{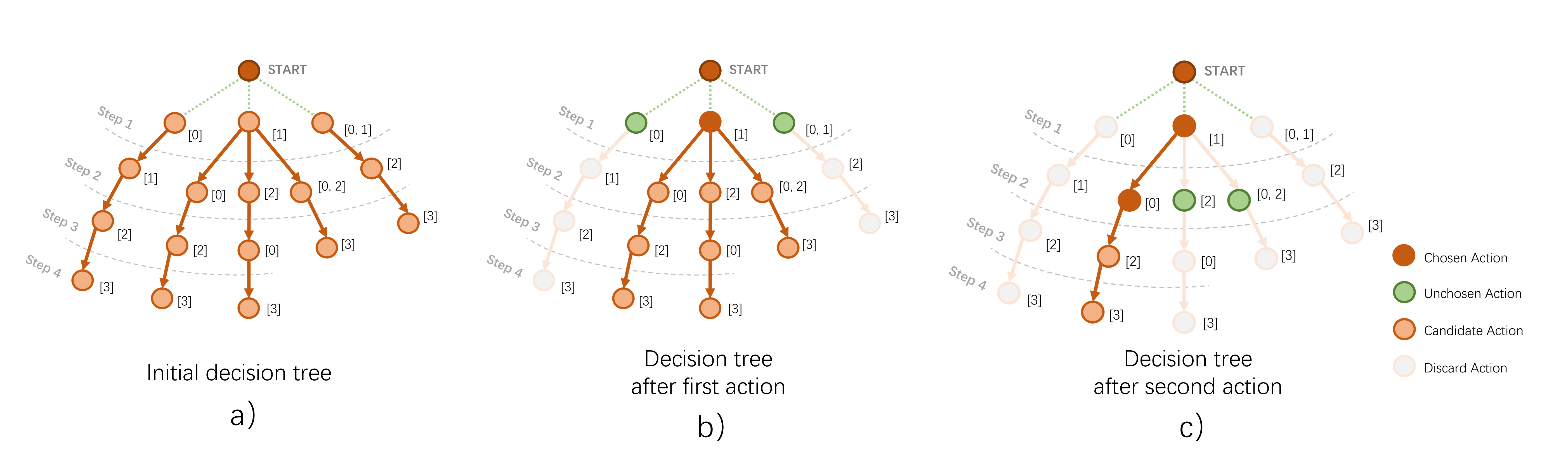} 
\caption {Visualization of the dynamic decision tree during evaluation.}
\label{fig:toy_2}
\end{figure*}

Consider a simplified example in Figure \ref{fig:toy_1}. A user wants to create a PowerPoint on popular movies and has four tools: [0] for presentation creation, [1] for retrieval of movie rank, [2] for detailed information gathering and [3] for slide generation. Here, [2] needs the parameters of [1], and [3] depends on [0] and [2].
Figure \ref{fig:toy_2} shows the evaluation process. 
The initial decision tree (Figure \ref{fig:toy_2} a ) based on tool dependencies has five candidate paths. When the agent calls Tool [1] first, we check its validity and prune relevant sub-trees (Figure \ref{fig:toy_2} b). The tool call [0] next leads to further pruning (Figure \ref{fig:toy_2} c). Finally, sequential matching verifies the remaining path.

\textbf{Accomplish Progress and Efficiency of Agents}
Based on the evaluation method, we introduce two fine-grained metrics to better observe model performance. 
Given the low success rate of complex tasks, we use an accomplish progress (AP) indicator to track incorrect steps in the decision tree.
To evaluate strategy efficiency, we compare the model tool-call steps to the optimal path, assign 1/0 tags and compute the agent's optimal path rate (OP Rate) for the whole test set.

The formal algorithm specifications are provided in the Appendix \ref{appendix-e}.

\subsection{Challenge 2: Critical Hidden Information}
Analysis of abundant user logs \cite{sharegpt} and previous research \cite{su2019improving} reveal that in sequential tasks, 80\% users modify or expand previous tasks, often omitting previous information. This requires agents to have robust contextual retrieval capabilities. To assess the model performance under such conditions, we designed a specific challenge.

\textbf{Test setup}:

1) During the test, the agents receive only historical user tasks (\(g\)) and LLM summaries (\(s\)) to prevent the influence of $P$. The test data contain 2-4 tasks; correct historical dialogues are provided to prevent cascading errors. Evaluation focuses solely on the final task.

2) All subsequent tasks lack necessary execution information, which is embedded within historical dialogues.

3) We propose three information hidden strategies: omitted information (\textbf{Omit}), referential information (\textbf{Reference}) and relying on distant info (\textbf{Long-Context}). We show demos in Appendix \ref{demo-3}. Common model failures include asking users to repeat the previous dialogue \ref{demo-20} and using incorrect tool parameters. 

For instance, in the first task, the user asks about the weather in New York. In the following task, "How about traffic conditions?", the location (New York) is \textbf{omitted}. In "What are its top recommended attractions", "its" is \textbf{reference} to New York. As the number of tasks increases, hidden information can be split by task, forming \textbf{long-context} scenarios. Moreover, with more tasks, the current one may need information from multiple previous tasks for decision making, increasing problem complexity.

\textbf{Evaluation Method}

We measure the impact of hidden information on model robustness via two dimensions:
Assessing the influence of different information hidden strategies on model accuracy.
Analyze how the number of tasks affects model accuracy.

Notably, while Challenge 1 and Challenge 2 may overlap within a single test case, they can be distinguished across the entire dataset, based on task length and information hidden types.

\subsection{Challenge 3: Variable Decision Trajectories}
The Action-Observation step ($\mathbb{P}$) is the core and unique factor in the Agent's interaction with the physical world. To verify its impact on agents, we design a new challenge by collecting diverse dynamic decision paths that induce agents to modify decisions.

Each user task pairs with a specific agent policy. For a given task length, we gather multi-task data covering all possible agent policy type switch combinations. For example, a two-task subset has 16 ($2^4$, 2 is the task number, 4 is the scale of policy type) continuous task combinations (e.g., $\{P_{single},P_{single}\}$, $\{P_{single},P_{multi}\}$), with 64 ($3^4$) and 256 ($4^4$) combinations in the three-task and four-task subsets, respectively. This uniformly distributed data enables an objective evaluation of the effects of different challenge policies on the model.

During testing, we inject the correct execution process $\mathbb{P}_{n - 1}$ into a test process of Challenge 2.

\textbf{Evaluation Method} 

We assess model robustness by analyzing the differences between the evaluation results of Challenge 2 and 3 ($R_{C2}$ and $R_{C3}$). First, we calculate the proportion of changed decisions. Then, we reveal what factors cause the fluctuations.

In partial testdata, the agent produced the right policy first, but answers incorrectly in the new test. Meanwhile, some previously incorrect answers improve. We use a table for illustration:

\begin{table}[ht]
\centering
\begin{tabular}{|c|c|c|}
\hline
\scriptsize \diagbox{Challenge 3}{Challenge 2} & \scriptsize Right & \scriptsize Wrong \\
\hline
\scriptsize Right & \scriptsize \textit{{RR}} &\scriptsize  \textit{{WR}} \\
\hline
\scriptsize Wrong &\scriptsize  \textit{{RW}} &\scriptsize \textit{{WW}} \\
\hline
\end{tabular}
\end{table}
Here, $\textit{RW}$ is the number of test data where the agent correctly answered in Challenge 2 but incorrectly in Challenge 3; $\textit{WR}$ is the opposite; $\textit{RR}$ is the number of correctly answered questions in both challenges. We assess the model's overall capability in the two tests as $Acc_2 = \textit{RR}/(\textit{RR}+\textit{WW}+\textit{WR}+\textit{RW})$.

To evaluate model robustness, we propose two metrics: one to measure the \textbf{change degree} for model stability and another for the \textbf{change direction} to find instability causes.

The Volatility Factor ($VF$) measures the change degree:
\begin{equation}
VF(R_{C2},R_{C3})=\frac{\textit{WR}+\textit{RW}}{\textit{WR}+\textit{RW}+\textit{RR}+\textit{WW}}
\end{equation}
Higher $VF$ indicates lower robustness. 

The Debiased Descent Direction ($DDD$) measures the change direction:
\begin{equation}
DDD(R_{C2},R_{C3})=\frac{\textit{RW}}{\textit{WR}}*\omega=\frac{\textit{RW}}{\textit{WR}}*\frac{1}{Acc_2}
\end{equation}

The $\textit{RW}/\textit{WR}$ ratio is the bidirectional conversion rate, showing accuracy change direction. Higher ratios mean stronger accuracy drop tendencies. We find that the task difficulty bias this ratio. For example, dealing with the harder questions of Challenge 2, the model produces fewer correct and more incorrect answers. This increases the probability of a small $\textit{RW}$ and a large $\textit{WR}$, resulting in a lower bidirectional conversion rate. Therefore, we divide it by $Acc_2$ to reduce bias, and obtain $DDD$ metric.

\begin{figure}[h]
  \centering
  \includegraphics[width=0.6\linewidth]{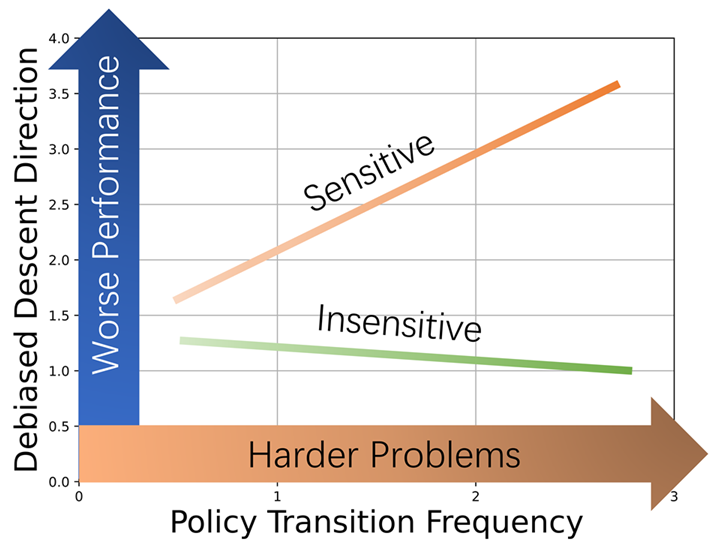} 
  \caption {Diagram: $DDD$ and $PTF$  Relationship.}
  \label{fig:toy}
\end{figure}

$DDD$ can cooperate with other factors to reveal influencing factors. In this work, we placed emphasis on observing the correlation between $PTF$ and $DDD$. We divide the test set by $PTF$ values, calculate $DDD$ for each subset. As shown in Figure \ref{fig:toy}, a positive correlation between higher $PTF$ and $DDD$ indicates that frequent policy switches reduce agent robustness; if $PTF$ increases while $DDD$ remains stable, the model is insensitive to policy changes.

\subsection{Test Data Collection}

In this study, data collection is crucial, as it must address three challenges concurrently. The data must satisfy the following conditions: 1) be high-quality multi-task data; 2) incorporate various tools; 3) ensure subsequent tasks are closely linked to previous ones; 4) exhibit broad diversity to cover all policy type switch combinations; 5) maintain an approximate uniform distribution of these factors in the benchmark.

To fulfill these requirements, we introduce a novel multi-role data generation framework, collaborating with human experts for data creation. The framework consists of five roles: User, Planner, AI Agent, Checker and Observer. For data diversity, we randomly choose one from HunYuan-Series of LLMs to generate user questions and provide tens of few-shot examples at the same time. Then we use HunYuan-TurboS to generate the execution process. The generation process is carefully controlled. We first sample the desired decision trajectory \(\mathbb{P}\), specify the policy type \({P}_i\) for each step to generate tasks, and randomly assign an information hiding method for subsequent tasks. Heuristic rules verify the data format during generation and the Checker role assesses its reasonableness.

The framework enriches human-inspired data, overcoming the diversity limitations of relying solely on humans. However, due to LLMs' limited generation accuracy, five LLM research experts refine the data, fixing errors and making user queries more natural and aligned with human-like distribution. Each annotator conducts self-checks after single-task and multi-round generations, followed by cross-checks among annotators and final review by a senior expert. This process increases annotation accuracy from about 60\% at initial generation to 86\% after two self-checks, 94\% after cross-checking and nearly 100\% after final verification. We detail the method in Appendix \ref{chap:4}.

\setlength{\intextsep}{1pt}    

\section{Experiments}

\begin{figure*}[h]
  \centering
  \includegraphics[width=0.9\linewidth]{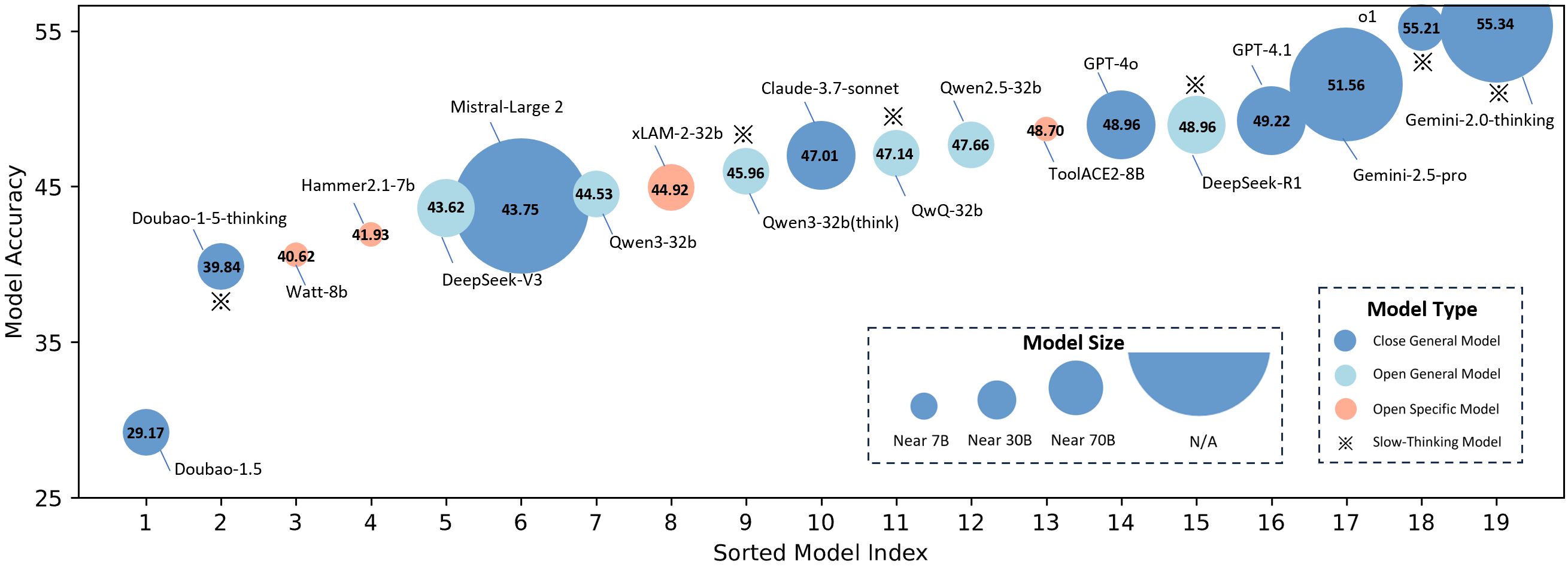} 
  \caption {Overall accuracy of agents on the multi-tasking. }
  \label{fig:exp_1}
\end{figure*}

\subsection{Overall Performance}

\begin{figure}[h]
  \centering
  \includegraphics[width=0.6\linewidth]{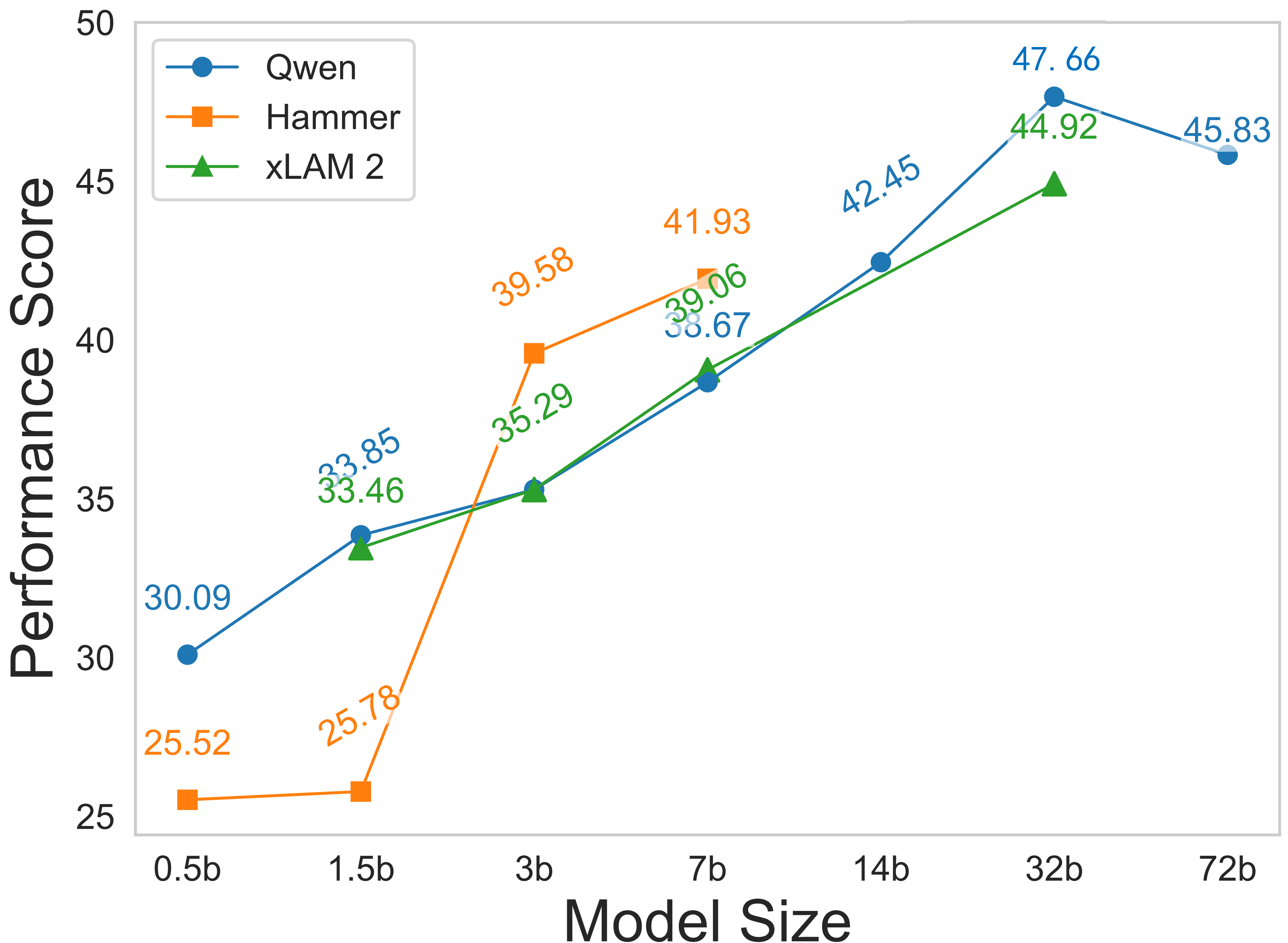} 
 \caption {Accuracy of 3 series of agents. }
  \label{fig:exp_2}
\end{figure}

The $C^3$-Bench has 256 single-task and 768 multi-task testdata. Challenge 1 contains 256 entries with 1-4 tasks, while Challenges 2 and 3 contain 768 multi-task data with 2-4 tasks.

We evaluated 49 state-of-the-art mainstream models. This paper analyzes 31 top-performing models; full results are in the Appendix\ref{appendix-x}. Models include \textbf{general fast-think}\cite{gpt4.1,claude,mistral-large,Doubao,liu2024deepseek,yang2024qwen2}, \textbf{slow-think}\cite{openai-o1,Gemini,doubao-think,qwen3,guo2025deepseek} and \textbf{specialized types}\cite{zeng2025toolace,liu2024apigen,lin2024hammer,shi2024direct}. For the Qwen3 series with hybrid thinking modes, we tested each mode separately. Models are open-source or closed-source. We use the recommended Function Call interface when available; otherwise, prompt.

This section reports the accuracy of 19 representative models. For each model series, we show only the best result. Testing follows Challenge 3's method.

Key findings of Figure \ref{fig:exp_1}:
1) The SOTA model scores 55.34, highlighting the limits of the LLM-based Agent.
2) In the top-5, 60\% are slow-think models, 80\% are closed-source general models.
3) SOTA open-source (DeepSeek-R1) and specialized (ToolACE2-8b) models lag behind the top models (e.g., Geimini-2.0-thinking, o1) but match other closed-source models. 
4) Generally, slow-think models outperform fast-think ones in the same series, except Qwen3-32b (hybrid), which shows fewer differences.

Figure \ref{fig:exp_2} shows the accuracy of 3 model series. This shows that the performance of the models correlates with the size of model. 

To understand the causes of model accuracy differences, we use fine-grained metrics to assess models based on three challenges in the following sections.

\linespread{0.5} 

\subsection{Challenge 1: Complex Tool Relationships}

\begin{table}
    \centering
    \caption{Agent accuracy, progress and efficiency in handling multi-tool calls tasks.}
    \label{tab:exp_1}
   \resizebox{1.0\columnwidth}{!}{ \begin{tabular}{l|c|ccc|cc}
    \toprule
    {\footnotesize Models} & 
    {\footnotesize Total} & 
    {\footnotesize $P_{\mathrm{multi}}^P$} & 
    {\footnotesize $P_{\mathrm{multi}}^S$} & 
    {\footnotesize $P_{\mathrm{multi}}^{S+P}$} &
    {\footnotesize AP} & 
    {\footnotesize OP Rate} \\
    \midrule

\scriptsize GPT-4.1 & \scriptsize \cellcolorheat{44.14}  & \scriptsize \cellcolorheat{58.60} & \scriptsize \cellcolorheat{25.00} & \scriptsize \cellcolorheat{20.24} & \scriptsize \cellcolorheat{46.06} & \scriptsize \cellcolorheat{43.57} \\
\scriptsize GPT-4o & \scriptsize \cellcolorheat{41.80}  & \scriptsize \cellcolorheat{53.50} & \scriptsize \cellcolorheat{31.25} & \scriptsize \cellcolorheat{21.43} & \scriptsize \cellcolorheat{45.06} & \scriptsize \cellcolorheat{41.08} \\
\scriptsize QwQ-32b & \scriptsize \cellcolorheat{41.41}  & \scriptsize \cellcolorheat{54.78} & \scriptsize \cellcolorheat{31.25} & \scriptsize \cellcolorheat{17.86} & \scriptsize \cellcolorheat{36.68} & \scriptsize \cellcolorheat{41.91} \\
\scriptsize DeepSeek-R1 & \scriptsize \cellcolorheat{41.02}  & \scriptsize \cellcolorheat{53.50} & \scriptsize \cellcolorheat{18.75} & \scriptsize \cellcolorheat{21.43} & \scriptsize \cellcolorheat{43.89} & \scriptsize \cellcolorheat{41.49} \\
\scriptsize Gemini-2.0-thinking & \scriptsize \cellcolorheat{40.23}  & \scriptsize \cellcolorheat{54.14} & \scriptsize \cellcolorheat{25.00} & \scriptsize \cellcolorheat{16.67} & \scriptsize \cellcolorheat{40.37} & \scriptsize \cellcolorheat{40.66} \\
\scriptsize o1 & \scriptsize \cellcolorheat{39.06}  & \scriptsize \cellcolorheat{50.96} & \scriptsize \cellcolorheat{12.50} & \scriptsize \cellcolorheat{21.43} & \scriptsize \cellcolorheat{37.35} & \scriptsize \cellcolorheat{39.83} \\
\scriptsize Claude-3.7 & \scriptsize \cellcolorheat{39.06}  & \scriptsize \cellcolorheat{43.57} & \scriptsize \cellcolorheat{62.50} & \scriptsize \cellcolorheat{25.00} & \scriptsize \cellcolorheat{63.65} & \scriptsize \cellcolorheat{0.00} \\
\scriptsize ToolACE2-8B & \scriptsize \cellcolorheat{37.89}  & \scriptsize \cellcolorheat{47.77} & \scriptsize \cellcolorheat{31.25} & \scriptsize \cellcolorheat{20.24} & \scriptsize \cellcolorheat{43.72} & \scriptsize \cellcolorheat{21.99} \\
\scriptsize Gemini-2.5-pro & \scriptsize \cellcolorheat{36.33}  & \scriptsize \cellcolorheat{49.04} & \scriptsize \cellcolorheat{25.00} & \scriptsize \cellcolorheat{14.29} & \scriptsize \cellcolorheat{39.87} & \scriptsize \cellcolorheat{32.37} \\
\scriptsize Mistral-Large-2 & \scriptsize \cellcolorheat{36.33}  & \scriptsize \cellcolorheat{47.77} & \scriptsize \cellcolorheat{25.00} & \scriptsize \cellcolorheat{16.67} & \scriptsize \cellcolorheat{40.87} & \scriptsize \cellcolorheat{35.27} \\
\scriptsize DeepSeek-V3 & \scriptsize \cellcolorheat{35.16}  & \scriptsize \cellcolorheat{48.41} & \scriptsize \cellcolorheat{25.00} & \scriptsize \cellcolorheat{11.90} & \scriptsize \cellcolorheat{27.47} & \scriptsize \cellcolorheat{34.85} \\
\scriptsize xLAM-2-32b & \scriptsize \cellcolorheat{34.77}  & \scriptsize \cellcolorheat{45.86} & \scriptsize \cellcolorheat{25.00} & \scriptsize \cellcolorheat{15.48} & \scriptsize \cellcolorheat{41.88} & \scriptsize \cellcolorheat{18.67} \\
\scriptsize Qwen2.5-32b & \scriptsize \cellcolorheat{32.03}  & \scriptsize \cellcolorheat{44.59} & \scriptsize \cellcolorheat{25.00} & \scriptsize \cellcolorheat{9.52} & \scriptsize \cellcolorheat{31.32} & \scriptsize \cellcolorheat{29.88} \\
\scriptsize Qwen3-32b(think) & \scriptsize \cellcolorheat{28.91}  & \scriptsize \cellcolorheat{42.04} & \scriptsize \cellcolorheat{12.50} & \scriptsize \cellcolorheat{7.14} & \scriptsize \cellcolorheat{29.48} & \scriptsize \cellcolorheat{28.63} \\
\scriptsize Watt-8b & \scriptsize \cellcolorheat{28.13}  & \scriptsize \cellcolorheat{44.59} & \scriptsize \cellcolorheat{6.25} & \scriptsize \cellcolorheat{1.19} & \scriptsize \cellcolorheat{21.94} & \scriptsize \cellcolorheat{29.46} \\
\scriptsize Doubao-1.5-pro & \scriptsize \cellcolorheat{24.61}  & \scriptsize \cellcolorheat{35.03} & \scriptsize \cellcolorheat{12.50} & \scriptsize \cellcolorheat{7.14} & \scriptsize \cellcolorheat{30.15} & \scriptsize \cellcolorheat{6.64} \\
\scriptsize Doubao-1-5-thinking & \scriptsize \cellcolorheat{22.66}  & \scriptsize \cellcolorheat{31.21} & \scriptsize \cellcolorheat{18.75} & \scriptsize \cellcolorheat{7.14} & \scriptsize \cellcolorheat{26.30} & \scriptsize \cellcolorheat{20.75} \\
\scriptsize Hammer2.1-7b & \scriptsize \cellcolorheat{21.88}  & \scriptsize \cellcolorheat{33.12} & \scriptsize \cellcolorheat{12.50} & \scriptsize \cellcolorheat{2.38} & \scriptsize \cellcolorheat{14.24} & \scriptsize \cellcolorheat{21.58} \\

    \bottomrule
    
    \end{tabular}}
\end{table}

Table \ref{tab:exp_1} shows 19 models' results on multi-tool tasks.  

\textbf{Complex Tool Handling}
Comparing with $P_{multi}^{S}$, model's correct rate is obvious lower in the $P_{multi}^{P},P_{multi}^{S + P}$ tasks, where the inter-tool dependencies cause \textbf{28.7\%} accuracy declines in the view of average accuracy of all models. This shows that complex tool relationships are a major obstacle for LLM-based Agents. Our proposed soft metric Accomplish Progress (AP) shows more information.

Notably, Gemini-2.0-thinking, Gemini-2.5-pro and o1 drop significantly in the rankings. Although perform well in total, 
they experienced an unexpected decline in accuracy due to unbalance policy development. We show a case in Appendix \ref{demo-19}.

\textbf{Efficiency of Policies}
We used the Optimal Path Rate (OP Rate) to assess the efficiency of model policies in $P_{multi}^{S}, P_{multi}^{P}$ tasks. Models like Claude-3.7 show a significant decrease in this metric, mainly because they cannot execute parallel invocations.

\subsection{Challenge 2: Critical Hidden Information}
Table \ref{tab:exp_2} reports the model test results under different information hiding strategies. We conclude that hidden information greatly affects model performance; even the top model only achieves an accuracy of \textbf{51.56}. Long-context information dependencies challenges the models most, which cause \textbf{11.0\%} more accuracy declines in the view of the average accuracy of all models. This indicates that hidden information is a critical challenge for LLM-based agents.

\begin{figure}[H] 
      \centering
      \includegraphics[width=0.8\linewidth]{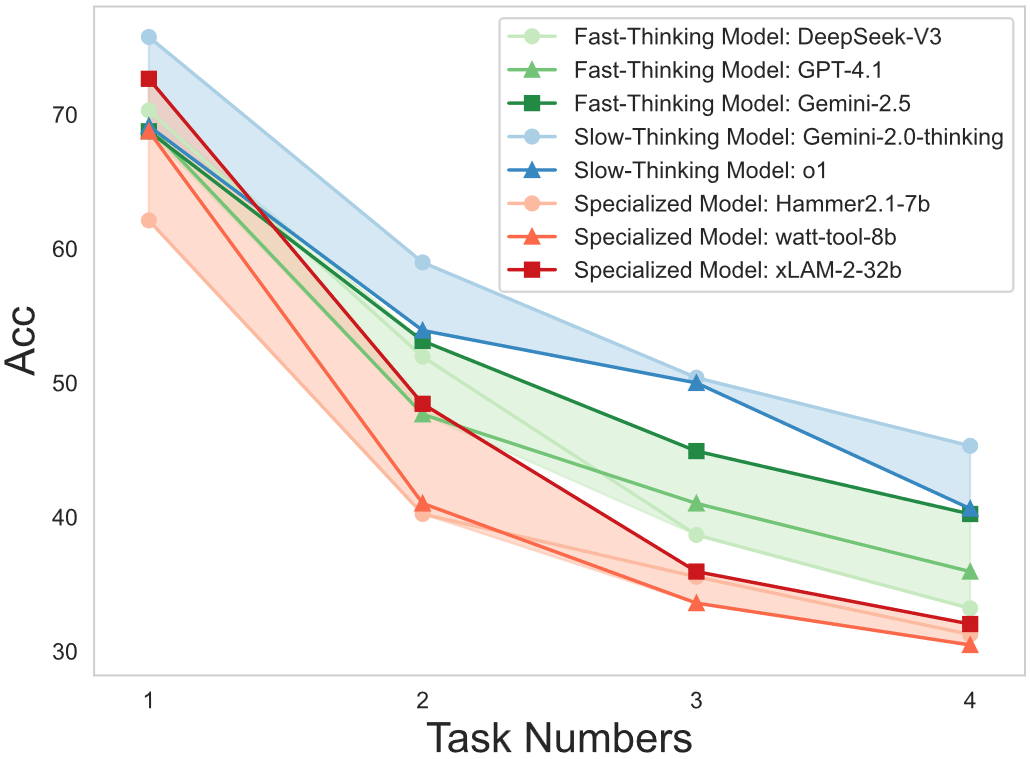} 
      \caption{Impact of task length on 3 categories agents.}
      \label{fig:exp_3}
    \end{figure}

In view of the average accuracy of all models, one more task causes \textbf{10.1\%} accuracy declines. Figure \ref{fig:exp_3} shows the performance of 3 model categories in 4 subsets with different task numbers. We specifically look at accuracy when the task number is 1 (a metric not included in the overall accuracy calculation). As the number of tasks increases, all models' accuracy decreases. Accuracy drops further when the task length is 3 or 4. 
This stems from the fact that long-text dependency primarily occurs in Rounds 3 and 4. Additionally, these two rounds may simultaneously depend on multi-round tasks.

    \begin{table}[H] 
      \caption{Impact of information hidden on agents. Here, `Ref',`Long' is short for reference and long-context.}
      \label{tab:exp_2}
      \scriptsize 
      \begin{tabular}{l|c|ccc}
        \hline
        \makecell{\textbf{Model}}  &\makecell{\textbf{Total}}  & \makecell{\textbf{Omit}} & \makecell{\textbf{Ref}} & \makecell{\textbf{Long}}\\
        \hline
        \scriptsize Gemini-2.0-thinking & \scriptsize \cellcolorheat{51.56}  & \scriptsize \cellcolorheat{54.44} & \scriptsize \cellcolorheat{55.00} & \scriptsize \cellcolorheat{41.34}  \\
        \scriptsize o1 & \scriptsize \cellcolorheat{48.18}  & \scriptsize \cellcolorheat{53.30} & \scriptsize \cellcolorheat{48.33} & \scriptsize \cellcolorheat{37.99}  \\
        \scriptsize QwQ-32b & \scriptsize \cellcolorheat{47.14}  & \scriptsize \cellcolorheat{46.13} & \scriptsize \cellcolorheat{53.33} & \scriptsize \cellcolorheat{40.78}  \\
        \scriptsize Qwen3-32b & \scriptsize \cellcolorheat{47.01}  & \scriptsize \cellcolorheat{49.00} & \scriptsize \cellcolorheat{49.58} & \scriptsize \cellcolorheat{39.66}  \\
        \scriptsize Gemini-2.5-pro & \scriptsize \cellcolorheat{46.09}  & \scriptsize \cellcolorheat{49.86} & \scriptsize \cellcolorheat{46.25} & \scriptsize \cellcolorheat{38.55}  \\
        \scriptsize ToolACE2-8B & \scriptsize \cellcolorheat{46.09}  & \scriptsize \cellcolorheat{47.28} & \scriptsize \cellcolorheat{50.83} & \scriptsize \cellcolorheat{37.43}  \\
        \scriptsize Qwen3-32b(think) & \scriptsize \cellcolorheat{43.62}  & \scriptsize \cellcolorheat{46.42} & \scriptsize \cellcolorheat{46.67} & \scriptsize \cellcolorheat{34.08}  \\
        \scriptsize GPT-4o & \scriptsize \cellcolorheat{43.49}  & \scriptsize \cellcolorheat{43.27} & \scriptsize \cellcolorheat{49.17} & \scriptsize \cellcolorheat{36.31}  \\
        \scriptsize DeepSeek-R1 & \scriptsize \cellcolorheat{42.32}  & \scriptsize \cellcolorheat{44.70} & \scriptsize \cellcolorheat{46.67} & \scriptsize \cellcolorheat{31.84}  \\
        \scriptsize GPT-4.1 & \scriptsize \cellcolorheat{41.54}  & \scriptsize \cellcolorheat{42.98} & \scriptsize \cellcolorheat{46.25} & \scriptsize \cellcolorheat{32.40}  \\
        \scriptsize DeepSeek-V3 & \scriptsize \cellcolorheat{41.28}  & \scriptsize \cellcolorheat{42.41} & \scriptsize \cellcolorheat{47.08} & \scriptsize \cellcolorheat{31.28}  \\
        \scriptsize xLAM-2-32b & \scriptsize \cellcolorheat{38.80}  & \scriptsize \cellcolorheat{39.83} & \scriptsize \cellcolorheat{42.92} & \scriptsize \cellcolorheat{31.28}  \\
        \scriptsize Doubao-1-5-thinking & \scriptsize \cellcolorheat{38.15}  & \scriptsize \cellcolorheat{38.40} & \scriptsize \cellcolorheat{40.83} & \scriptsize \cellcolorheat{34.08}  \\
        \scriptsize Claude-3.7 & \scriptsize \cellcolorheat{37.63}  & \scriptsize \cellcolorheat{34.38} & \scriptsize \cellcolorheat{45.42} & \scriptsize \cellcolorheat{33.52}  \\
        \scriptsize Qwen2.5-32b & \scriptsize \cellcolorheat{36.59}  & \scriptsize \cellcolorheat{36.68} & \scriptsize \cellcolorheat{43.33} & \scriptsize \cellcolorheat{27.37}  \\
        \scriptsize Hammer2.1-7b & \scriptsize \cellcolorheat{35.68}  & \scriptsize \cellcolorheat{41.26} & \scriptsize \cellcolorheat{34.58} & \scriptsize \cellcolorheat{26.26}  \\
        \scriptsize Mistral-Large-2 & \scriptsize \cellcolorheat{35.29}  & \scriptsize \cellcolorheat{35.24} & \scriptsize \cellcolorheat{39.17} & \scriptsize \cellcolorheat{30.17}  \\
        \scriptsize Watt-8b & \scriptsize \cellcolorheat{35.03}  & \scriptsize \cellcolorheat{41.55} & \scriptsize \cellcolorheat{32.08} & \scriptsize \cellcolorheat{26.26}  \\
        \scriptsize Doubao-1.5-pro & \scriptsize \cellcolorheat{25.91}  & \scriptsize \cellcolorheat{25.79} & \scriptsize \cellcolorheat{28.33} & \scriptsize \cellcolorheat{22.91}  \\
        \hline
      \end{tabular}
    \end{table}

  \setlength{\columnsep}{30pt}  

When handling independent tasks (task number = 1), specialized models like xLAM-2-32b outperform most general models. But in multi-tasking, specialized models experience the sharpest accuracy drop, followed by general fast-thinking models. This pattern shows the limitations of both model types in managing multiple tasks.

\subsection{Challenge 3: Variable Decision Trajectories}

\begin{figure}

  \centering
  \includegraphics[width=0.8\linewidth]{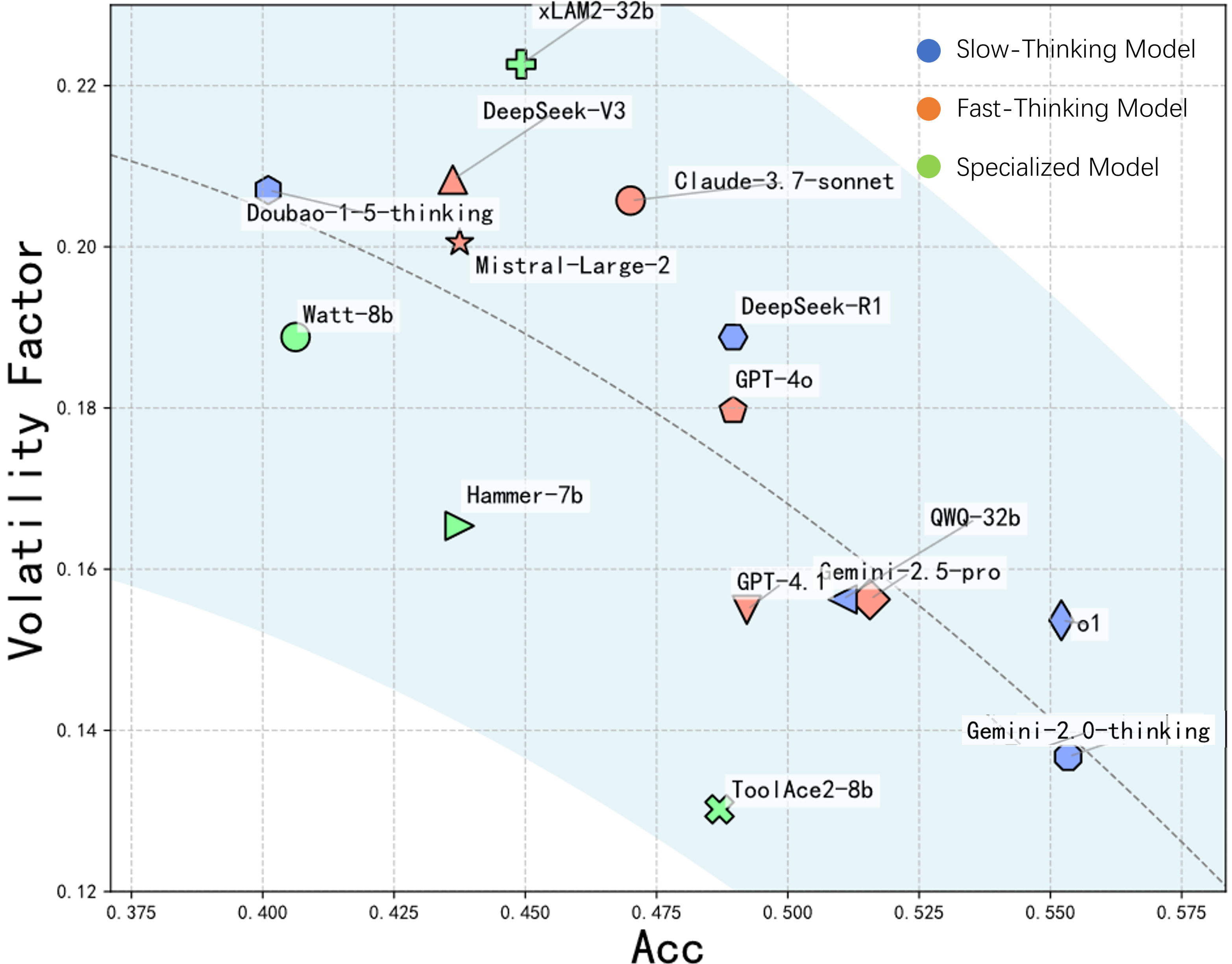} 
   \caption {The relation between accuracy and $VF$ of agents.}
  \label{fig:exp_4}

\end{figure}

Figure \ref{fig:exp_4} shows the relationship between the model accuracy and the volatility ($VF$). Models with lower volatility tend to have higher accuracy. 
This pattern indicates from another perspective that the decision-making process largely determines the model's capability. The less a model is influenced by previous decision-making processes, the better its performance tends to be. We further use the $DDD$ metric to reveal how the decision-making process affects stability ($VF$).
\setlength{\intextsep}{5pt}    
\setlength{\columnsep}{10pt}    

\setlength{\intextsep}{3pt}    

\begin{figure}[h]
  \centering
  \includegraphics[width=1.0\linewidth]{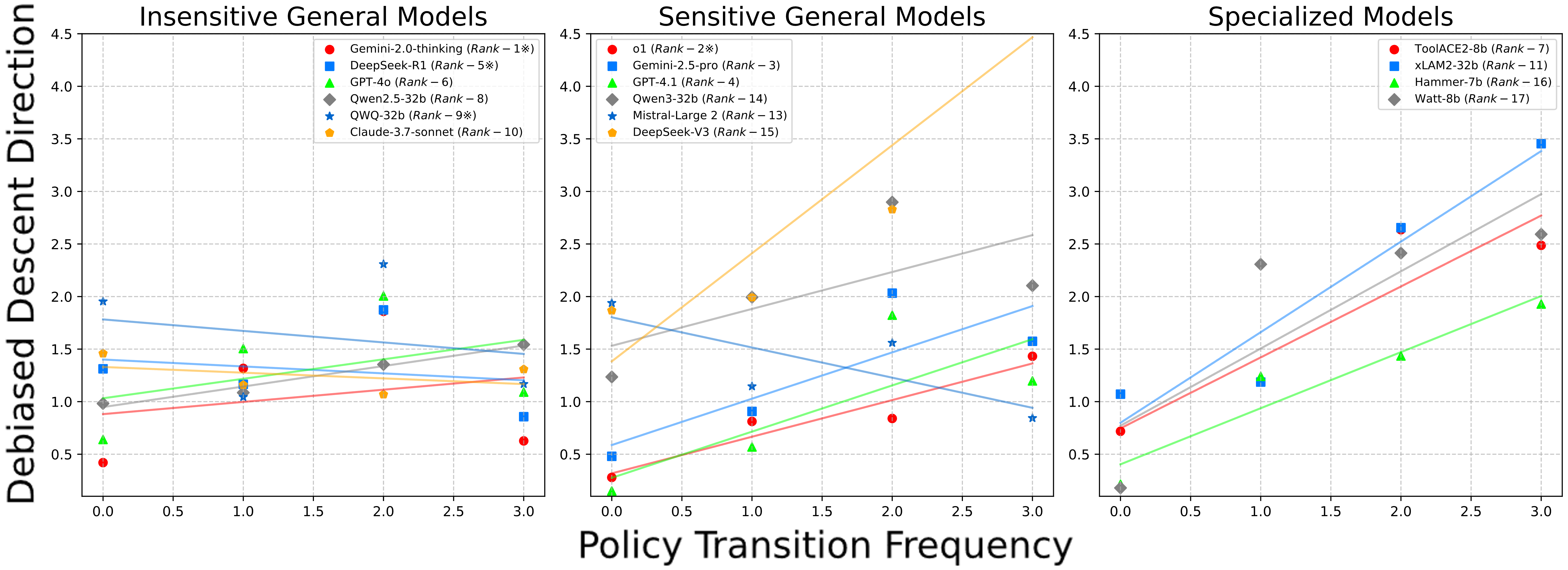} 
  \caption {Chart of agent's $DDD$ values across Various $PTF$ subsets.}
  \label{fig:exp_5}
\end{figure}

Figure \ref{fig:exp_5} illustrates the trends of $DDD$ and $PTF$ metrics in 3 model groups.
All but 6 models are sensitive to the $PTF$, especially the specialized models. Sensitivity to $PTF$ does not directly affect the accuracy of the model, but helps explain the behavior of model. If a model has high $VF$ and its $DDD$ is sensitive to $PTF$, we can provide the following `diagnosis': \textbf{The agent suffers damage in $VF$ and the `murderer' is $PTF$}.

\begin{figure}[h]
  \centering
  \includegraphics[width=1.0\linewidth]{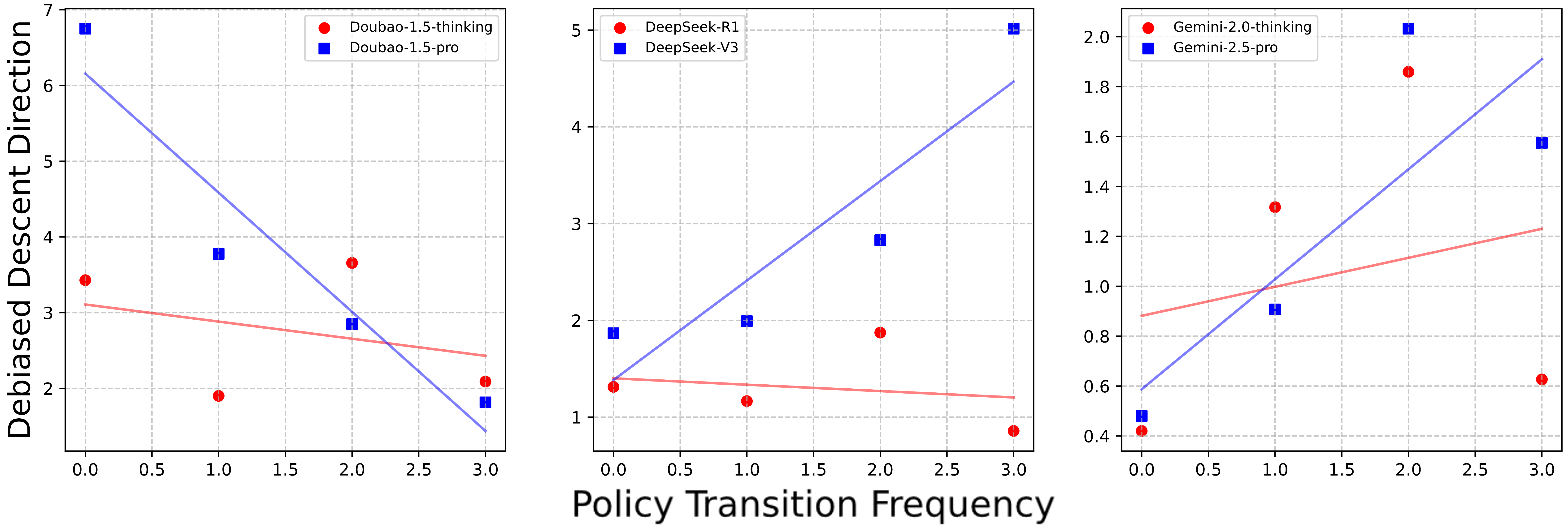} 
  \caption {Compare the fast- and slow-thinking agent in $DDD$ and $PTF$ correlation.}
  \label{fig:exp_6}
\end{figure}

Figure \ref{fig:exp_6} compares fast- and slow-thinking models within the same series. Slow-thinking models are less sensitive to $PTF$ changes. This suggests that more thinking time reduces the impact of frequent policy changes. 

Although the slow-thinking mode shows promising results in multiple aspects, it is \textbf{not a one-size-fits-all solution}. Deepseek-R1 outperforms Deepseek-V3 by 5\% in accuracy and their average output lengths for a single task are 603.4 and 110.8 tokens, respectively. To boost efficiency and cut costs, many models skip the cot processes. Thus, the accuracy-efficiency trade-off decides whether to adopt the slow-thinking mode. As discussed in Appendix \ref{appendix-xx}, \textbf{increasing the model size} is another possible solution.

\setlength{\intextsep}{30pt}    
\setlength{\intextsep}{1pt}    

\section{Related Work}

\subsection{Evaluation of LLMs}
Recent benchmarks evaluate LLM-based agents from various angles. Some studies assess generalizability across scenarios \cite{liembodied,trivedi2024appworld,liuagentbench}. Others explore how tool diversity impacts performance by gathering numerous tools \cite{duanytool,qintoolllm,ye2024tooleyes,li2023api}. Certain research focuses on domain-specific agents \cite{zhuang2023toolqa,guo2024ctooleval,xietravelplanner}. Some comprehensively assess multiple agent abilities \cite{shen2024taskbench,chen2024t,huang2024planning}, while others address specific issues like the illusion problem \cite{patil2023gorilla} and multi-step execution \cite{shen2024shortcutsbench,yao2024tau}.

BFCL V3 \cite{gorilla-openfunctions-v3} is closest to our work, but it only covers a small part of policy-switching space. Our benchmark simulates all possible policy transitions and all tasks in our data are related, unlike BFCL V3's independent tasks. 

WorfBench and TaskBench \cite{qiao2024benchmarking,shen2024taskbench} use graph-based methods for multi-tool evaluation, but only compute path-annotation similarity, while our benchmark can determine correctness and optimal probability (Table \ref{tab:benchmark_comparison}). 

Notably, we \textbf{increase policy types by 50\%, policy combination types by 150\% and the proportion of related tasks by 540\%} compared to the best benchmarks. 
\subsection{LLM-as-Agent}
Automating user missions is a key LLM research area. General LLMs integrate it into multi-task learning \cite{achiam2023gpt,sun2024hunyuan,yang2024qwen2} and there are many specialized agents exist.

Agent research varies in approach. Some endow agents with self-reflection \cite{xu2024rethinking,qiao2024autoact,zhang2024agent}, others introduce heuristic frameworks \cite{zhang2024xlam,han2024ibsen,islam2024mapcoder} and some strengthen core skills \cite{shi2024direct,schick2023toolformer} or generate diverse data \cite{functionary-medium,lin2024hammer}. Our benchmark features a novel data generation framework specifying desired agent actions, simulates real-world scenarios, evaluates core abilities and tests both general and specialized LLMs. 

\section{Conclusion}
$C^3$-Bench poses three challenges for agents based on five factors of multi-task execution and introduces several fine-grained metrics to evaluate model capabilities and analyze model thinking. Experiments on 49 mainstream models show that current SOTA models struggle with complex tasks, and closed-source models as well as slow-thinking models perform better. Research also finds that tool reliance, long-context dependencies, task length and dynamic policies are key factors the challenging model capabilities. These findings will improve the model interpretability and promote the performance of the agent.

\section{Limitations}
$C^3$-Bench uses expert annotations to ensure data quality, diversity and alignment with human distribution. However, this limits the scaling potential of the data size. Otherwise, the dual objectives of maintaining data quality and traversing all policy transition types concurrently limit the feasible length of tasks.

Despite this, experimental results reveal clear trends in model performance, leading to sound conclusions. Additionally, the poor performance of agents reduces the need for a larger test set.

\bibliography{custom}
\newpage

\appendix

\section{Methods}
\subsection{Test Data Collection}\label{chap:4}
\subsubsection{Diverse Toolset Construction}\label{appendix-a}

We build our toolset by leveraging tool descriptions from public APIs, following the approach introduced by ToolAlpaca. This API repository encompasses a collection of 400 tools, which can be grouped into 50 distinct categories, totaling 1600 tools.

In contrast to ToolAlpaca, our method incorporates three key strategies to enhance tool accuracy and diversify parameters. First, we use large language models (LLMs) such as HunYuan-TurboS to refine tool descriptions. This step is crucial because many generated tools lack constraint parameters. For example, a tool designed to query the weather in Spanish-speaking regions might not specify Spain as a valid location in any of its three main functions, making it impossible to validate the query location. Second, we broaden the range of parameter types to include complex data structures like enumerations, arrays, and objects. This adjustment better reflects real-world usage scenarios. Finally, a panel of five LLM agent experts reviews the generated tools to ensure their accuracy and the diversity of their parameters.

To create multi-task test data and thoroughly explore the task-switching space, we've introduced a novel data generation framework. This section explains the framework and how we constructed the benchmark. Subsection \ref{chap:4.1} outlines the five roles within the framework and their interaction mechanisms. Subsection \ref{chap:4.2} details how these roles collaborate to complete a single task, including defining task types and establishing dependencies between tasks. Subsection \ref{chap:4.3} expands our scope from generating individual tasks to creating a comprehensive test dataset with multiple related tasks. We then thoroughly explore the task-switching space to build the entire benchmark. Additionally, Appendices \ref{appendix-a} and \ref{appendix-b} provide details on how we collected the tools and the distribution of the test set.

\begin{figure*}[t]
  \centering
  \includegraphics[width=1.0\linewidth]{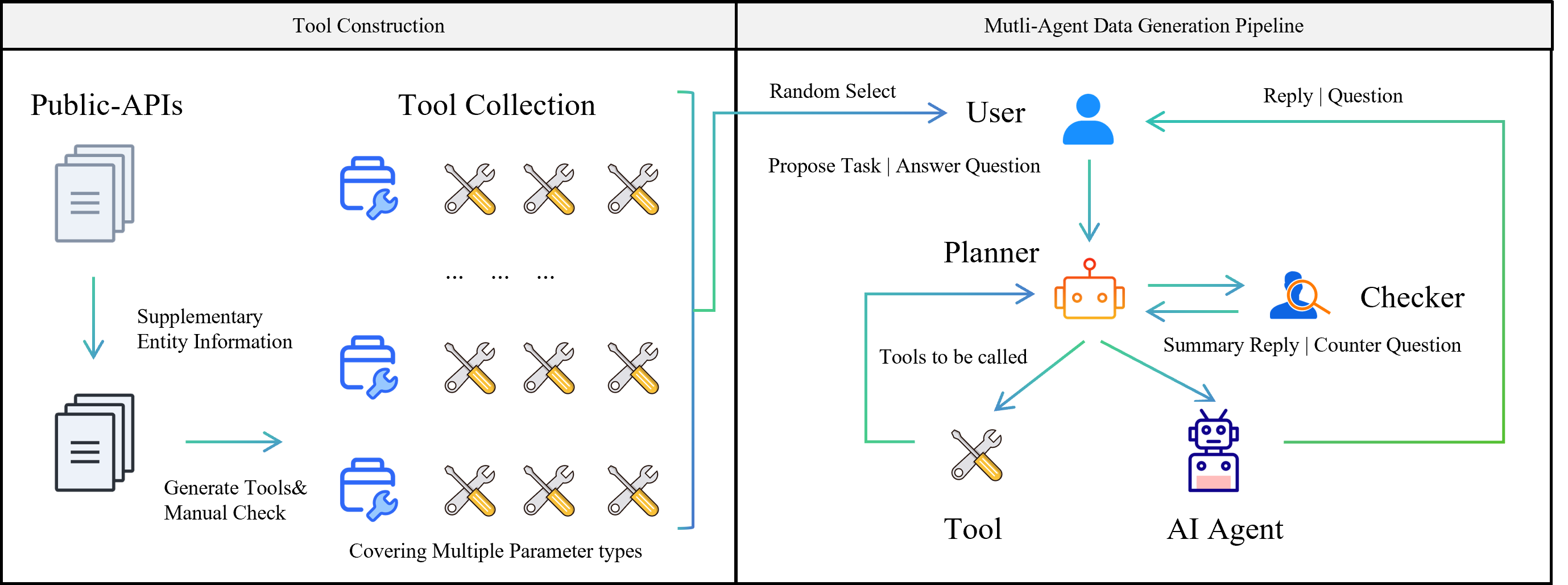} 
  \caption {The multi-agent framework.}
  \label{fig:five_role}
\end{figure*}

\begin{figure*}[h]
  \centering
  \includegraphics[width=1.0\linewidth]{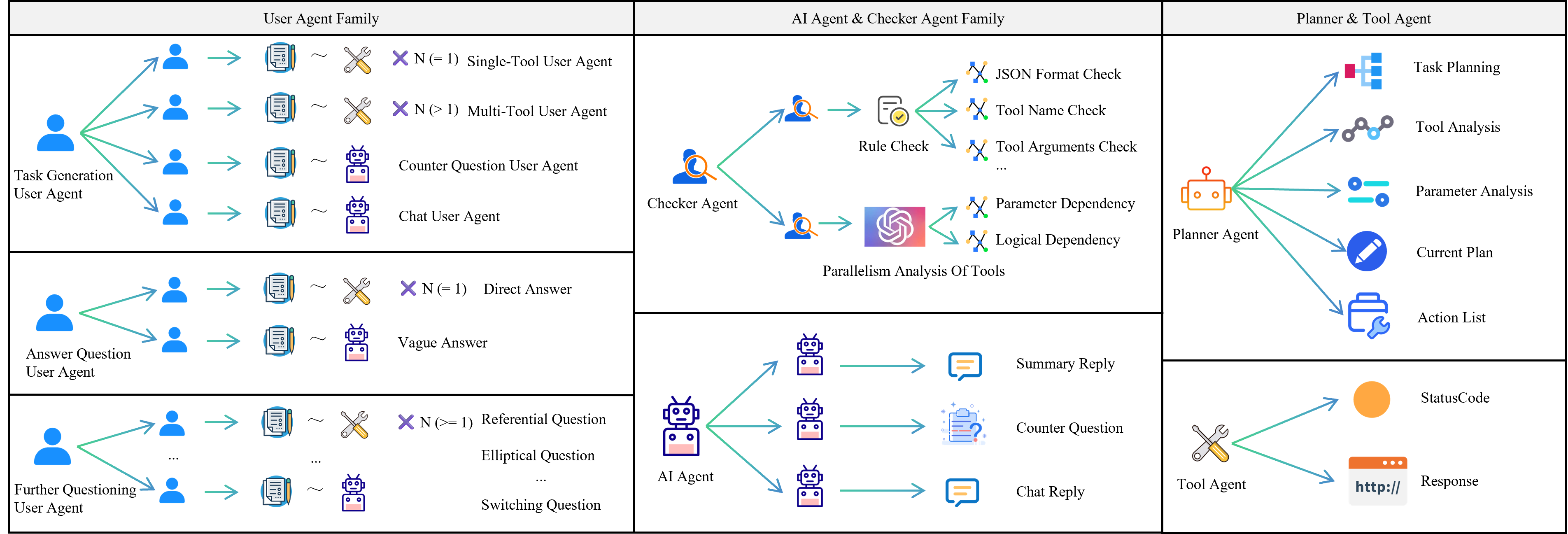} 
  \caption {The functions of each role}
  \label{fig:func}
\end{figure*}

\subsubsection{Data Generation Framework}\label{chap:4.1}

Our approach uses five agents to generate multi-task test data. Although this process is simulated with a single LLM, each agent is assigned a distinct role and specific tasks, denoted as $\mathbf{R}$. These roles are the User ($R_{user}$), Planner ($R_{planner}$), AI ($R_{AI}$), Checker ($R_{checker}$), and Tool ($R_{tool}$). The Planner plays a central role by analyzing tasks, planning tool invocation paths, and determining action types. Figure \ref{fig:five_role} illustrates the interactions among these roles, while Figure \ref{fig:func} details their specific functions.

In this framework, only $R_{AI}$ communicates directly with $R_{user}$, and $R_{planner}$ receives instructions from $R_{user}$. When $R_{planner}$ initiates a single task ($P_{single}$) or a multi-task sequence ($P_{multi}$), $R_{tool}$ simulates tool feedback. For clarification tasks ($P_{clarify}$) or chat interactions ($AP{chat}$), $R_{AI}$ asks about tool parameters or summarizes responses. $R_{checker}$ ensures the accuracy of $R_{planner}$'s plans by checking their format and sequence, though it is only involved in the data generation phase. It's important to note that $R_{user}$ has different responsibilities at different stages: $R_{user}^{Q}$ is tasked with generating new tasks, while $R_{user}^{A}$ answers questions from $R_{AI}$.

The prompts for these roles are provided in Appendix \ref{appendix-c}.

\subsubsection{Generate Single task}\label{chap:4.2}

We'll first explain how to construct a single task using our multi-agent framework.

The process begins with generating user tasks. To do this, we first select a relevant tool list for the task at hand.

To achieve the desired task type, we incorporate a predefined action type $A_i$ into the role prompt $R_{user}^{Q}$.

For generating related tasks, we first create several candidate tasks. These candidates are then refined by experts to produce the final successive task. We classify task relationships into three types: implicit understanding, ellipsis, and long-term memory. By inserting these relationship types into $R_{user}^{Q}$, we generate three candidate tasks. $R_{user}^{Q}$ also includes previous user-AI dialogues. Finally, we manually select and refine these candidates to obtain the final task.

Once the user tasks are defined, we utilize the five roles described earlier to carry out the entire execution process.

When generating the first-round user tasks, apply diversity constraints to guide the model to generate five potential questions. Then, human experts select and revise the best one from these questions to determine it as the formal question. The specific diversity constraints are as follows:

\begin{figure*}[htbp]
\begin{tcolorbox}[title={Diversity constraints.}]

1. The user's tasks should use different types of sentence structures: imperative, declarative, interrogative, etc.
\medskip

2. The user's tasks should include different tones: colloquial, formal, polite, direct, etc.
\medskip

3. Ensure that the length of the user's tasks varies, gradually increasing from short to long.
\medskip

4. Ensure that the user's tasks involve different themes/instances, different scenarios, and different roles.
\medskip

5. Extract common entities that appear in all descriptions from the [Tool List] and ensure that these entities appear in the user's tasks.
\medskip

6. Do not explicitly specify the tool $\{\{\{tool\}\}\}$ in the user's tasks.
\end{tcolorbox}
\caption{Diversity constraints.}
\label{P-task}
\end{figure*}

When generating subsequent tasks, randomly select one from 22 few-shot examples to guide the model to generate a subsequent task. Similarly, human experts revise the generated questions. Here are the details of two few-shot examples:

\begin{figure*}[htbp]
\begin{tcolorbox}[title={Few-shot Examples.}] 

\textbf{1st Few-shot}
\medskip

When you ask questions, you need to refer to the content of the Agent's response. For example:
\medskip

   Use positive ordinal numbers to refer to the Agent's responses.
   \medskip

   Example:
   \medskip

   \textbf{User}: What signs can indicate that high-priced stocks are about to fall? (The task you proposed in the previous round)
   \medskip

   \textbf{Agent}: 1. Downward signals in technical analysis 2. Market sentiment and trading volume analysis ... (The Agent's response in the previous round)
   \medskip

   \textbf{User}: Please elaborate on the second point. (Proposed new task)
   \medskip

   Explanation: "Second point" refers to "2. Market sentiment and trading volume analysis" in the Agent's response, meaning "Please elaborate on the second point (referring to 2. Market sentiment and trading volume analysis)."

\medskip
\medskip

\textbf{2nd Few-shot}
\medskip

When you ask questions, you need to refer to the content of the Agent's response. For example:
\medskip

   Use reverse ordinal reference to the Agent's response.
   \medskip

   Example:
   \medskip

   \textbf{User}: Fun basketball plan (the task you proposed in the previous round)
   \medskip

   \textbf{Agent}: High-level volume stagnation, high-level large fluctuations ... Bearish rebound pattern (the response from the Agent in the previous round)
   \medskip

  \textbf{ User}: Elaborate on the last point (proposing a new task)
   \medskip

   Explanation: "The last point" is a reverse ordinal reference to the "bearish rebound pattern" in the Agent's response, meaning "elaborate on the last point (referring to the bearish rebound pattern)."

\end{tcolorbox}
\caption{Few-shot Examples.}
\label{P-task}
\end{figure*}

\subsubsection{Construct the Whole Benchmark}\label{chap:4.3}

In Subsection \ref{chap:4.2}, we established the process for generating a specific type of task and creating related tasks. We now apply this methodology to construct a benchmark that fully captures the diversity of task switching in test data. Our goal is to explore the entire task-switching space for a predefined number of tasks.

First, we identify all possible combinations of action types for a given number of tasks, denoted as $\mathbb{P} = \{\mathbf{P}_1^1, \mathbf{P}_1^2, ..., \mathbf{P}_N^{4^N}\}$. Here, $\mathbf{P}_i^j$ represents the $j$-th combination for $i$ tasks. For $i$ tasks, there are $4^i$ possible combinations.

Next, we generate test data independently for each action-type combination. If a combination consists of $N$ elements, we use the generation framework $N$ times to construct the corresponding test data. It's important to note that the outputs from both $R_{tool}$ and $R_{AI}$ are critical pieces of information that we provide to the agents during the testing process.

\subsubsection{Data Verification Mechanism}

Our verification process builds on the heuristic checks outlined in Section 2.5 and includes multiple layers of manual reviews:

1. \textbf{Single-Round Self-Check}:
   - Annotators ensure that the data adheres to guidelines, checking for response relevance, factual accuracy, appropriate language use, diverse questions, whether the strategies match the questions, and whether the questions effectively hide necessary information.
   - Internal tools are used to flag issues such as hallucinations or inconsistent reasoning.

2. \textbf{Multi-Round Self-Check}:
   - The focus here is on dialogue coherence, the ability to retain long-term context, and task completion.
   - The data is cross-referenced with gold-standard examples to ensure consistency.

3. \textbf{Cross-Check Phase}:
   - Pairs of annotators review each other's work independently.
   - Any discrepancies (found in more than 5\% of cases) are escalated to a senior expert for resolution.
   - Metrics such as inter-annotator agreement (IAA) and the F1 score for task completion are tracked.

4. \textbf{Expert Verification}:
   - A senior expert reviews all escalated cases and randomly samples 10\% of the approved annotations.
   - Corrective feedback loops are implemented to address recurring issues, such as insufficient evidence citation.

\textbf{Accuracy Metrics}:
- \textbf{Baseline}: 62\% (raw generation without checks)
- \textbf{After Single-Round Self-Check}: 78\%
- \textbf{After Multi-Round Self-Check}: 86\%
- \textbf{After Cross-Check}: 94\%
- \textbf{After Expert Verification}: 100.0\%

This rigorous verification process ensures the reliability of the dataset for various downstream applications, including model training, benchmarking, and academic publication.

\subsection{Details of Proposed Evaluation Method}\label{appendix-e}

1. Initialize graph G, indegree table, visitation table, current path, and all paths.

2. Perform topological sorting and depth-first traversal based on parallel combination and permutation:
   2.1 For each search, identify all nodes with an indegree of 0 and generate all possible combinations based on the number of nodes. Since nodes with an indegree of 0 are independent, they can be combined in any way. When a combination has more than one node, it means these nodes can be called in parallel. This approach allows our algorithm to enumerate all possible paths, including those with parallel and serial-parallel calls, unlike naive topological sorting methods that are limited to serial calls.
   
   2.2 Traverse each combination, add it to the current path, and update the indegree and visitation tables.
   
   2.3 Continue the depth-first traversal until the number of nodes in the path matches the number of nodes in the annotated answer. Once this is achieved, the path is added to the collection of all paths.
   
   2.4 Repeat the above steps until the entire graph has been traversed.

3. Based on the path length, classify the paths into optimal and suboptimal categories.

\section{More Experiments}
\subsection{Hyperparameter Settings}
To further enhance the reproducibility of our dataset, we hereby introduce the hyperparameter settings used during model inference. Specifically:

For \textbf{closed-source models}, we adopted the default hyperparameters from the official website without making any changes to hyperparameters such as \texttt{temperature}, \texttt{top-p}, and \texttt{top-k}.

For \textbf{open-source models}, we used the default hyperparameters in Hugging Face's \texttt{model.generate} method, with only the output \texttt{max\_new\_tokens} set to 512. The version of Hugging Face used was 4.51.0, and no other modifications were made.

If the model does not provide function call protocal, we employ the following prompt to test it:

\begin{figure*}[htbp]
\label{propmt}
\begin{tcolorbox}[title={Test Prompt}]
You are an expert in function composition. You will be given a question and a set of possible functions. Based on the question, you need to make one or more function/tool calls to achieve the purpose.
If none of the functions can be used, please directly reply to the user in natural language, starting with "Assistant:".
If the given question lacks the parameters required by the function, please ask the user for the necessary information in natural language, starting with "Assistant:".
If the result of the call is already sufficient to answer the user's question, please summarize the historical results and reply to the user in natural language, starting with "Assistant:".
You should only return function calls in the tool call section. If you decide to make any function calls, you must format them as <tool\_calls>[{"name": "funcn\_ame1", "arguments": {"argument1": "value1", "argument2": "value2"}},...]</tool\_calls>. You should not include any other text in your reply. The following is a list of functions you can call, in JSON format.
\medskip

\{\{\{tools\}\}\}
\medskip

If you decide to return function calls, please format them as <tool\_alls>[{"name": "func\_name1", "arguments": {"argument1": "value1", "argument2": "value2"}},...]</tool\_calls>, without including any other text.
Otherwise, please refer to the three cases mentioned at the beginning and reply starting with "Assistant:".
\medskip

Current time: \{\{\{env\_info\}\}\}

\end{tcolorbox}
\caption{Test Prompt}
\label{P-task1}
\end{figure*}

\subsection{The Detail Results of all Agents}\label{appendix-x}
\begin{figure*}[h]
  \centering
  \includegraphics[width=1.0\linewidth]{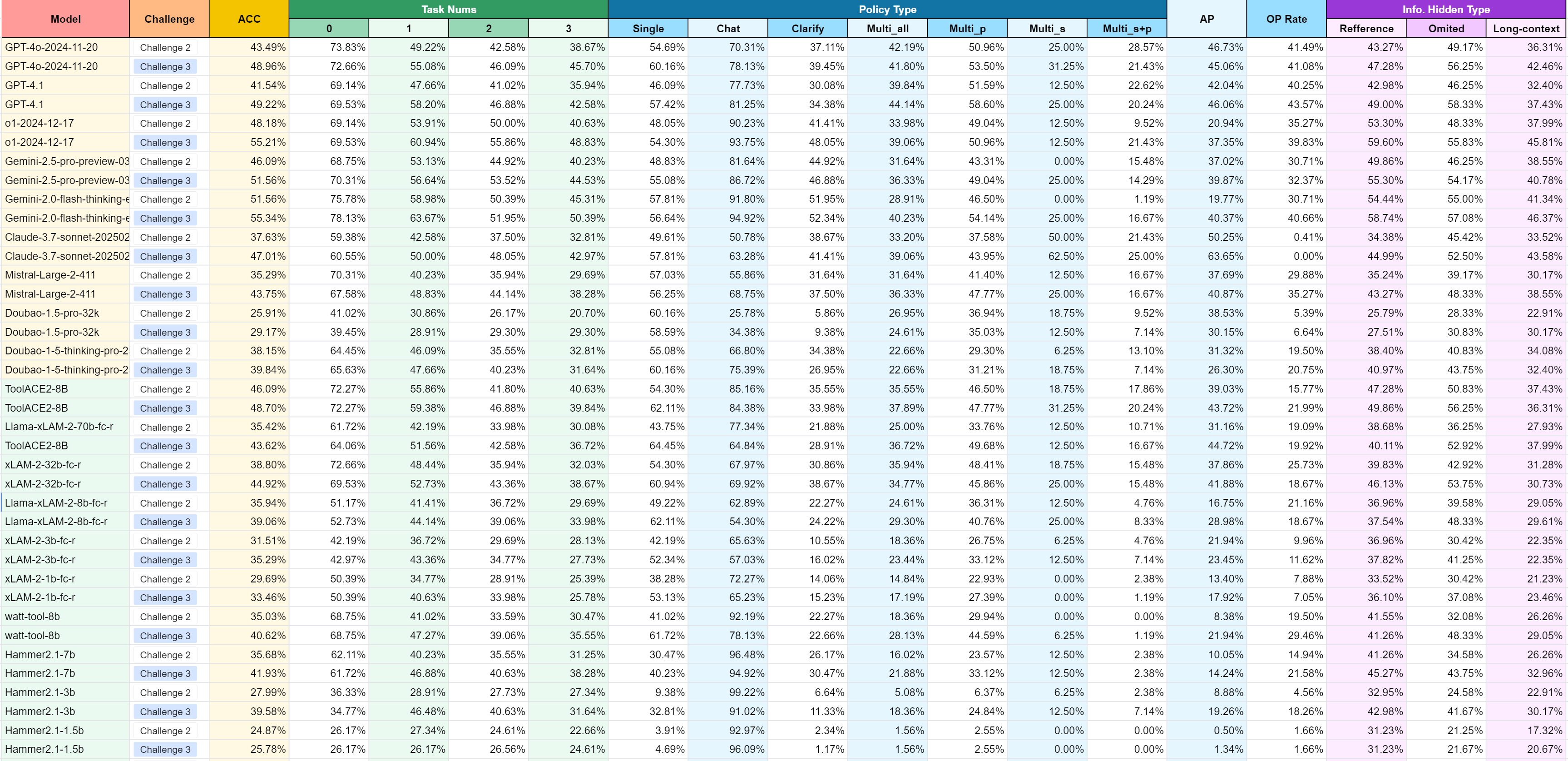} 
  \caption {Screenshot of Leaderboard.}
  \label{fig:all}
\end{figure*}

We provide the performance data of 49 models in Challenge 2 and Challenge 3 via an attached XLSX file, enabling readers to independently explore additional insights. Figure \ref{fig:all} displays a screenshot of this file, showcasing the structured results for easy reference.

\subsection{Error Types of the Agents}\label{appendix-f}
Beyond the analytical perspectives in the main text, we conduct a detailed analysis of agent error types. Errors are categorized into two main classes: tool errors and parameter errors. Parameter errors are further subdivided into three subcategories: parameter name hallucinations, parameter value hallucinations, and parameter value errors. Table \ref{tab:apdx_1} outlines this classification system. 

Notably, stronger agents exhibit a lower incidence of tool errors. While parameter name hallucinations occur less frequently, they are both severe and widespread. The most common parameter error involves incorrect extraction of parameter values. The table includes heatmap shading to visually highlight error proportions across different agents, facilitating quick comparative analysis.

\begin{table}[h]
    \centering
    \caption{The distribution of agent errors. Here, `Hallu.' is short for hallucination. }
    \resizebox{1.0\columnwidth}{!}{\begin{tabular}{l|c|c|c|c}
        \hline
        \multirow{3}{*}{\textbf{Agent}} & \multirow{3}{*}{\textbf{\makecell{Tool \\ Errors (\%)}}} & \multicolumn{3}{c}{\textbf{Parameter Errors (\%)}} \\
        \cline{3-5}
        & & \makecell{Name \\ Hallu.} & \makecell{Value \\  Hallu. } & \makecell{Value \\ Err} \\
        \hline
        o1-2024-12-17 & 83.33 & \cellcolorheatblueblue{0.24} & \cellcolorheatblueblue{5.07} & \cellcolorheatblueblue{11.36} \\
GPT-4o-2024-11-20 & 75.87 & \cellcolorheatblueblue{0.20} & \cellcolorheatblueblue{8.05} & \cellcolorheatblueblue{15.49} \\
Gemini-1.5-Pro-002 & 85.15 & \cellcolorheatblueblue{0.19} & \cellcolorheatblueblue{3.34} & \cellcolorheatblueblue{11.32} \\
Qwen2.5-72b-Instruct & 80.90 & \cellcolorheatblueblue{0.37} & \cellcolorheatblueblue{6.31} & \cellcolorheatblueblue{12.43} \\
ToolACE-8B & 90.56 & \cellcolorheatblueblue{0.17} & \cellcolorheatblueblue{1.75} & \cellcolorheatblueblue{7.52} \\
Mistral-Large-2411 & 78.19 & \cellcolorheatblueblue{0.35} & \cellcolorheatblueblue{6.46} & \cellcolorheatblueblue{15.01} \\
watt-tool-8b & 90.68 & \cellcolorheatblueblue{0.17} & \cellcolorheatblueblue{3.63} & \cellcolorheatblueblue{5.53} \\
GLM-4-9B-Chat & 92.99 & \cellcolorheatblueblue{0.15} & \cellcolorheatblueblue{2.99} & \cellcolorheatblueblue{3.88} \\
DeepSeek-R1 & 95.77 & \cellcolorheatblueblue{0.00} & \cellcolorheatblueblue{2.11} & \cellcolorheatblueblue{2.11} \\
doubao-1.5-pro-32k & 82.35 & \cellcolorheatblueblue{0.28} & \cellcolorheatblueblue{10.69} & \cellcolorheatblueblue{6.67} \\
xLAM-7b-fc-r & 96.36 & \cellcolorheatblueblue{0.27} & \cellcolorheatblueblue{1.35} & \cellcolorheatblueblue{1.89} \\
gorilla-openfunctions-v2 & 98.83 & \cellcolorheatblueblue{0.00} & \cellcolorheatblueblue{0.26} & \cellcolorheatblueblue{0.90} \\
DeepSeek-V3 & 96.57 & \cellcolorheatblueblue{0.00} & \cellcolorheatblueblue{0.90} & \cellcolorheatblueblue{2.53} \\
Llama-3.3-70B-Instruct & 90.53 & \cellcolorheatblueblue{0.33} & \cellcolorheatblueblue{2.45} & \cellcolorheatblueblue{6.69} \\
\hline
    \end{tabular}}
\label{tab:apdx_1}
\end{table}

\subsection{Relation between Scale and Sensitivity to $PTF$}\label{appendix-xx}

\begin{figure}[h]
  \centering
  \includegraphics[width=0.80\linewidth]{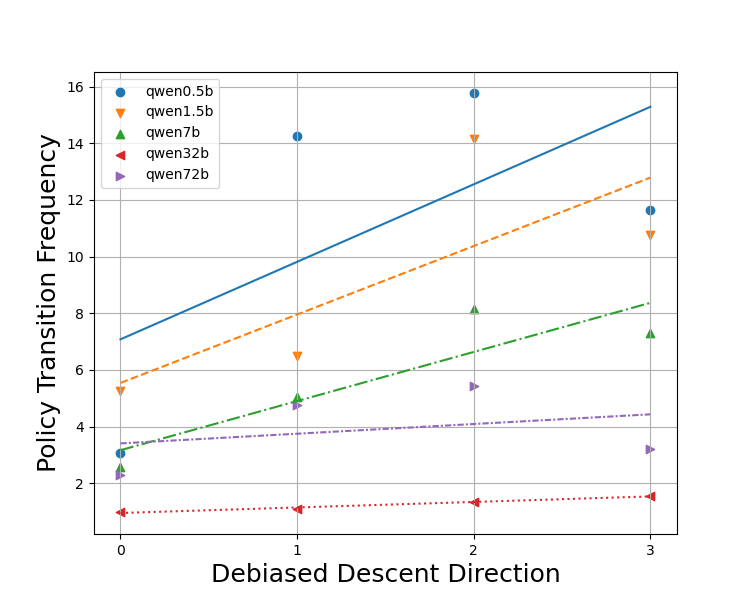} 
  \caption {$DDD$ of Qwen2.5 series corresponding to $PTF$.}
  \label{fig:qwen}
\end{figure}

By comparing test results across the Qwen2.5 series models (Figure \ref{fig:qwen}), a clear trend emerges: larger models demonstrate reduced sensitivity to changes in $PTF$. This indicates that model scale correlates with greater robustness against fluctuations in prompt-to-function mapping, a critical insight for model selection and optimization.

\section{Analysis of Benchmark}
\subsection{Analysis of the Test Data}\label{appendix-b}
Table \ref{tab:benchmark_comparison} highlights key differentiators between our benchmark and others. Our benchmark uniquely supports multi-tasking (MutTsk$^{*}$) with 100\% coverage of related tasks (RelTsk$^{\dag}$) and full policy switching type diversity ($\rm PST_4^{\ddag}$) across all mission types, including single-task, chat, clarification, and mixed serial-parallel multi-tasks. In contrast, other benchmarks like BFCL and ToolBench show limited or no support for these advanced features.

\begin{table*}[h]
\centering
\scriptsize  
\begin{tabular}{lccccccccc}
\hline 
\multirow{2}{*}{\makecell[c]{\textbf{Benchmark}}}  & \multirow{2}{*}{\makecell[c]{\textbf{MutTsk$^{*}$}}} & \multirow{2}{*}{\makecell[cc]{\textbf{Rate of}\\ \textbf{RelTsk$^{\dag}$}} } & \multirow{2}{*}{\makecell[cc]{\textbf{PST}$_4^{\ddag}$}} & \multicolumn{6}{c}{\textbf{Mission-Types}} \\
\cline{5-10} 
 &  & &  & \makecell{$A_{single}$} & \makecell{$A_{chat}$} & \makecell{$A_{clarity}$} & \makecell{$A_{multi}^S$} & \makecell{$A_{multi}^P$} & \makecell{$A_{multi}^{S+P}$}  \\
\hline
\textbf{Ours} & \textcolor{teal}{\Checkmark}  &\textbf{\underline{100} } &\textbf{\underline{100}}  & \textcolor{teal}{\Checkmark} & \textcolor{teal}{\Checkmark} & \textcolor{teal}{\Checkmark} & \textcolor{teal}{\Checkmark} & \textcolor{teal}{\Checkmark} & \textcolor{teal}{\Checkmark}\\
\rowcolor{gray!20}
BFCL v3 & \textcolor{teal}{\Checkmark} & \textbf{15.7} & \textbf{39.7}  & \textcolor{teal}{\Checkmark}  & \textcolor{teal}{\Checkmark}  & \textcolor{teal}{\Checkmark}  & \textcolor{red}{\XSolidBrush} & \textcolor{teal}{\Checkmark} & \textcolor{red}{\XSolidBrush} \\

BFCL v1& \textcolor{red}{\XSolidBrush} & 0.0 & 0.9  & \textcolor{teal}{\Checkmark}  &  \textcolor{teal}{\Checkmark}  & \textcolor{red}{\XSolidBrush} & \textcolor{red}{\XSolidBrush} & \textcolor{teal}{\Checkmark} & \textcolor{red}{\XSolidBrush}  \\
\rowcolor{gray!20}
BFCL v2 & \textcolor{red}{\XSolidBrush} & 0.0 & 0.9  & \textcolor{teal}{\Checkmark}  &  \textcolor{teal}{\Checkmark}  & \textcolor{red}{\XSolidBrush} & \textcolor{red}{\XSolidBrush} & \textcolor{teal}{\Checkmark} & \textcolor{red}{\XSolidBrush}  \\
ToolBench & \textcolor{red}{\XSolidBrush} & 0.0 &  0.0 &\textcolor{teal}{\Checkmark}  & \textcolor{red}{\XSolidBrush} & \textcolor{red}{\XSolidBrush} & \textcolor{teal}{\Checkmark}   & \textcolor{red}{\XSolidBrush}&  \textcolor{red}{\XSolidBrush}\\
\rowcolor{gray!20}
AnyToolBench & \textcolor{red}{\XSolidBrush} & 0.0 &  0.0 &\textcolor{teal}{\Checkmark}  & \textcolor{red}{\XSolidBrush} & \textcolor{red}{\XSolidBrush} & \textcolor{teal}{\Checkmark}   & \textcolor{red}{\XSolidBrush}&  \textcolor{red}{\XSolidBrush}\\
$\tau$-bench  & \textcolor{red}{\XSolidBrush} & 0.0 &  0.0 &\textcolor{teal}{\Checkmark}  & \textcolor{red}{\XSolidBrush} & \textcolor{red}{\XSolidBrush} & \textcolor{teal}{\Checkmark}   & \textcolor{red}{\XSolidBrush}&  \textcolor{red}{\XSolidBrush}\\
\rowcolor{gray!20}
T-EVAL & \textcolor{red}{\XSolidBrush} & 0.0 &  0.0 &\textcolor{teal}{\Checkmark}  & \textcolor{red}{\XSolidBrush} & \textcolor{red}{\XSolidBrush} & \textcolor{teal}{\Checkmark}   & \textcolor{red}{\XSolidBrush}&  \textcolor{red}{\XSolidBrush}\\
UltraTool & \textcolor{red}{\XSolidBrush} & 0.0 &  0.0 &\textcolor{teal}{\Checkmark}  & \textcolor{red}{\XSolidBrush} & \textcolor{red}{\XSolidBrush} & \textcolor{teal}{\Checkmark}   & \textcolor{red}{\XSolidBrush}&  \textcolor{red}{\XSolidBrush}\\
\rowcolor{gray!20}
\hline
\end{tabular}

\caption{Comparative Analysis of the $C^3$-Bench against other benchmarks in the field. The symbol `*' indicates Multi-Task, while `\dag' denotes Related Tasks. Moreover, in the task length of 1-4, the policy switching type Scale ( $\rm PST_4$ ) represents the proportion of policy switching types to all possible types.}
\label{tab:benchmark_comparison}
\end{table*}

\begin{figure}[h]
  \centering
  \includegraphics[width=0.90\linewidth]{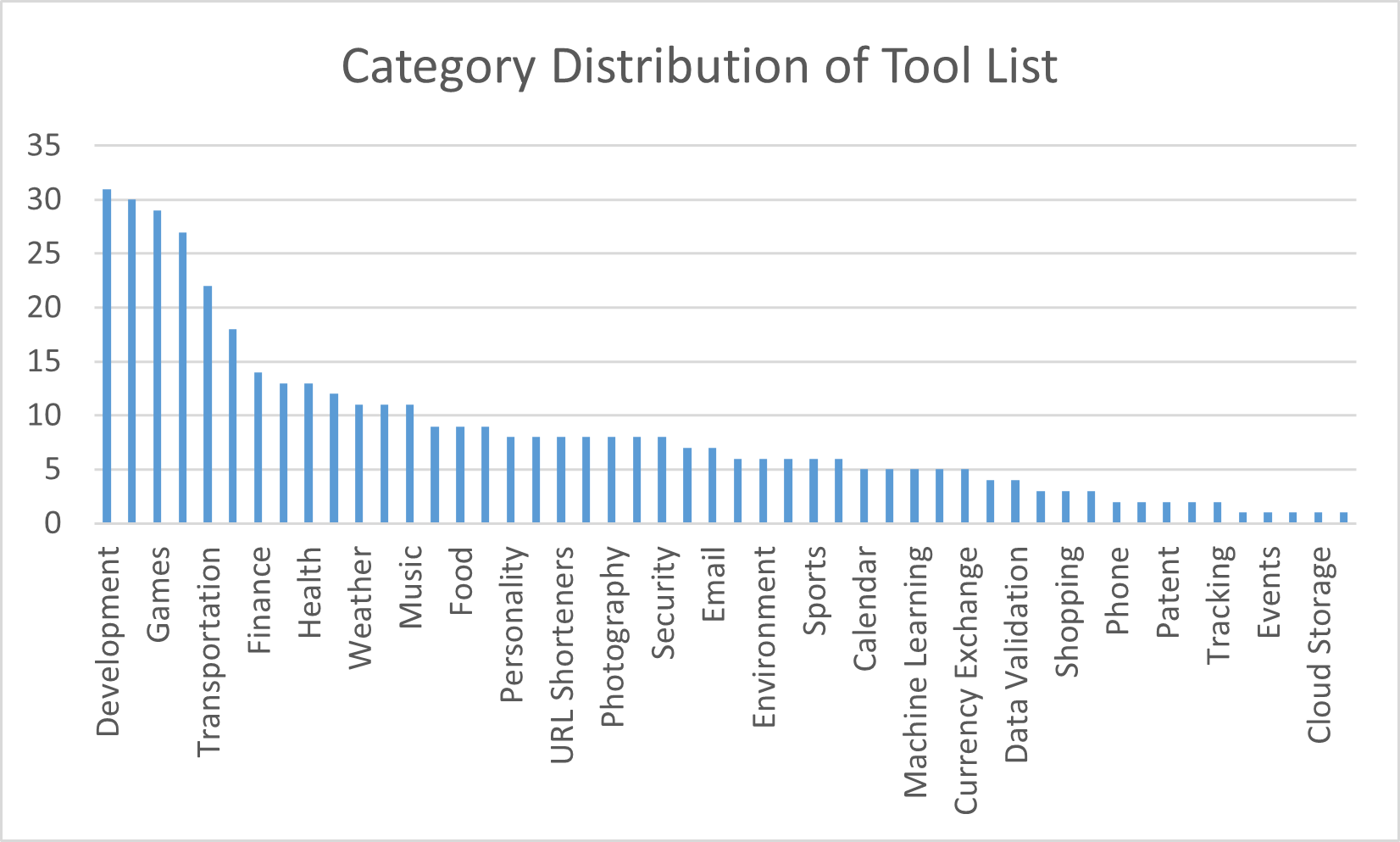} 
  \caption {Category distribution of tools.}
  \label{fig:apdx_1}
\end{figure}

\begin{figure}[h]
  \centering
  \includegraphics[width=0.90\linewidth]{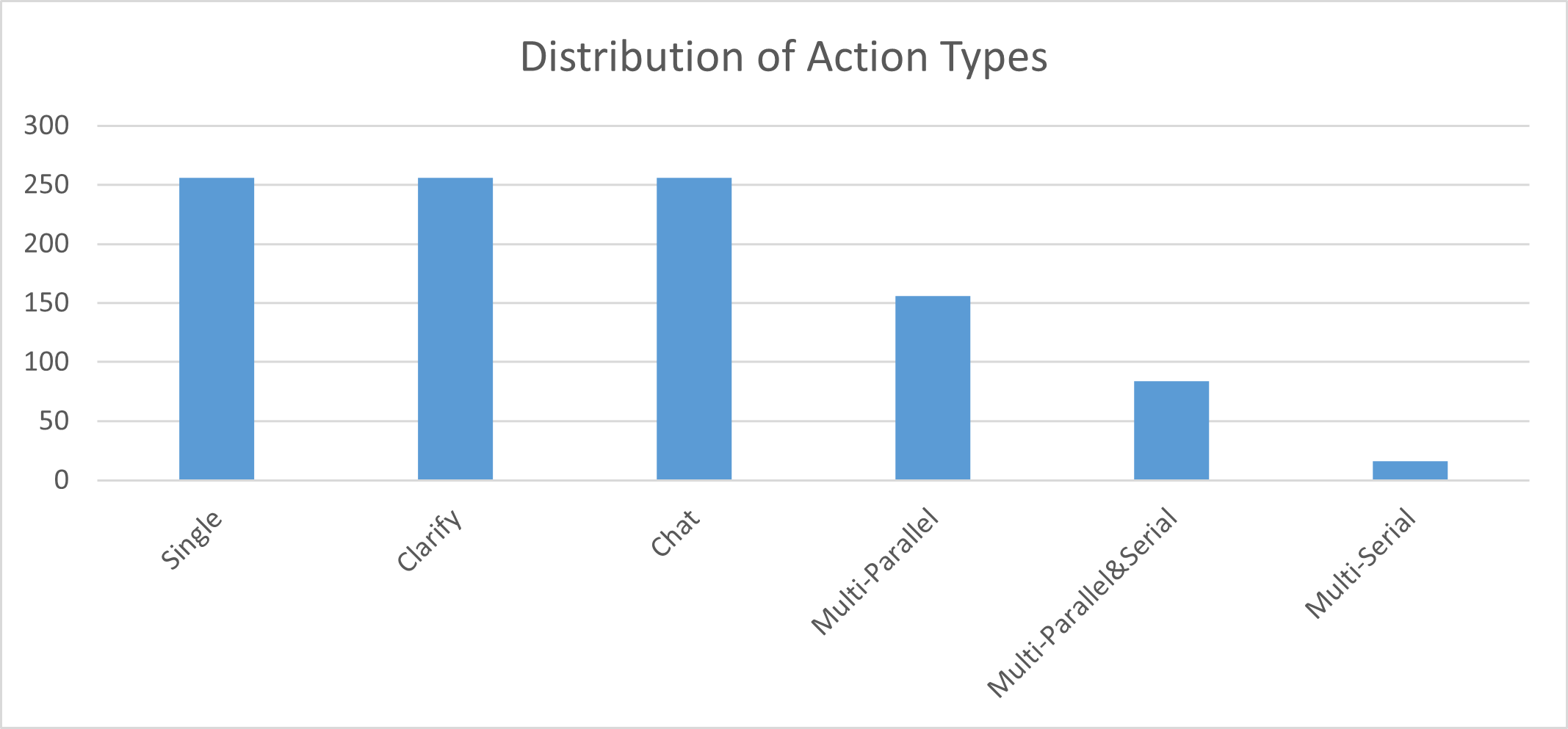} 
  \caption {Distribution of action-types.}
  \label{fig:apdx_2}
\end{figure}

\begin{figure}[h]
  \centering
  \includegraphics[width=0.7\linewidth]{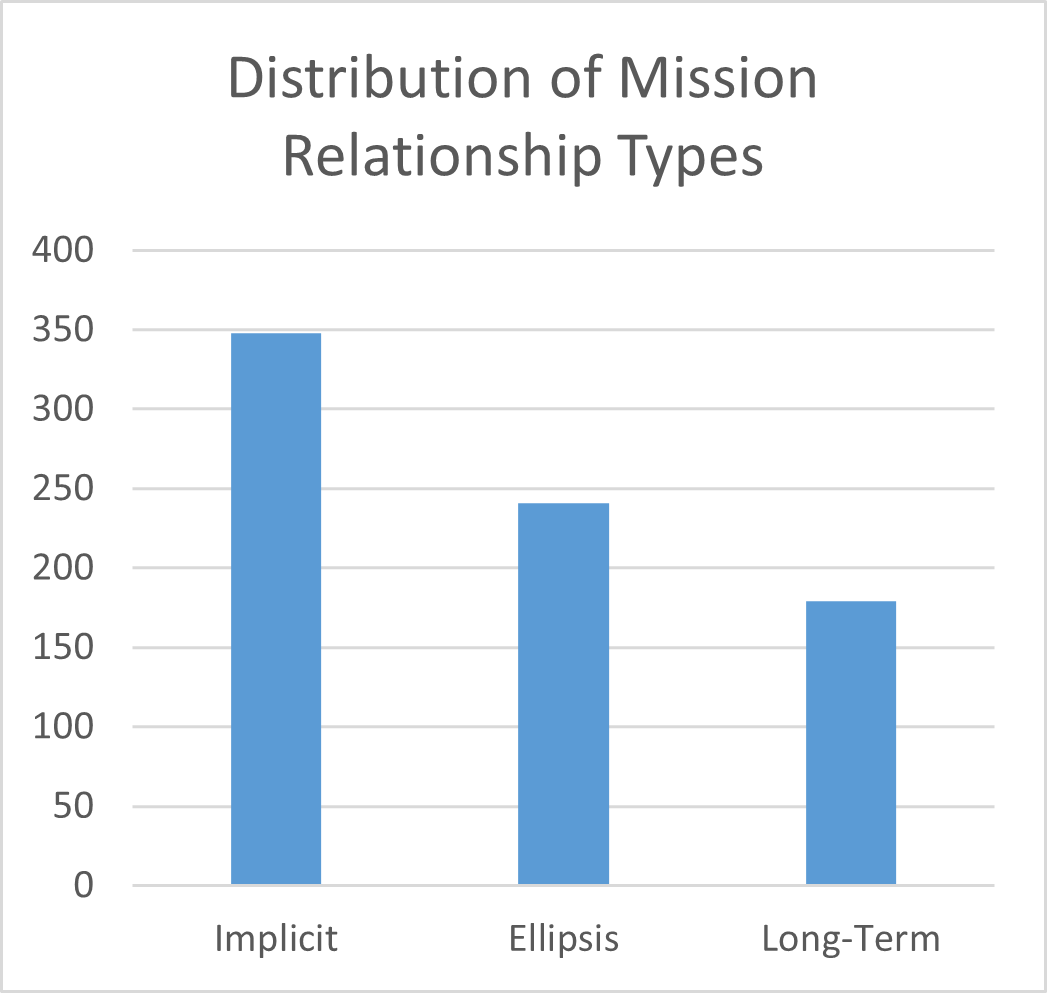} 
  \caption {Distribution of three task relationship types.}
  \label{fig:apdx_3}
\end{figure}

Figures \ref{fig:apdx_1}, \ref{fig:apdx_2}, and \ref{fig:apdx_3} offer three distinct perspectives on our dataset. Figure \ref{fig:apdx_1} shows the category distribution of tools, highlighting the diversity of functional domains. Figure \ref{fig:apdx_2} illustrates the distribution of action types, demonstrating balanced coverage of different task execution modes. Figure \ref{fig:apdx_3} depicts the distribution of three task relationship types, emphasizing the benchmark's focus on realistic task dependencies.

\subsection{Data Examples}
\begin{figure*}[h]
  \centering
  \includegraphics[width=1.0\linewidth]{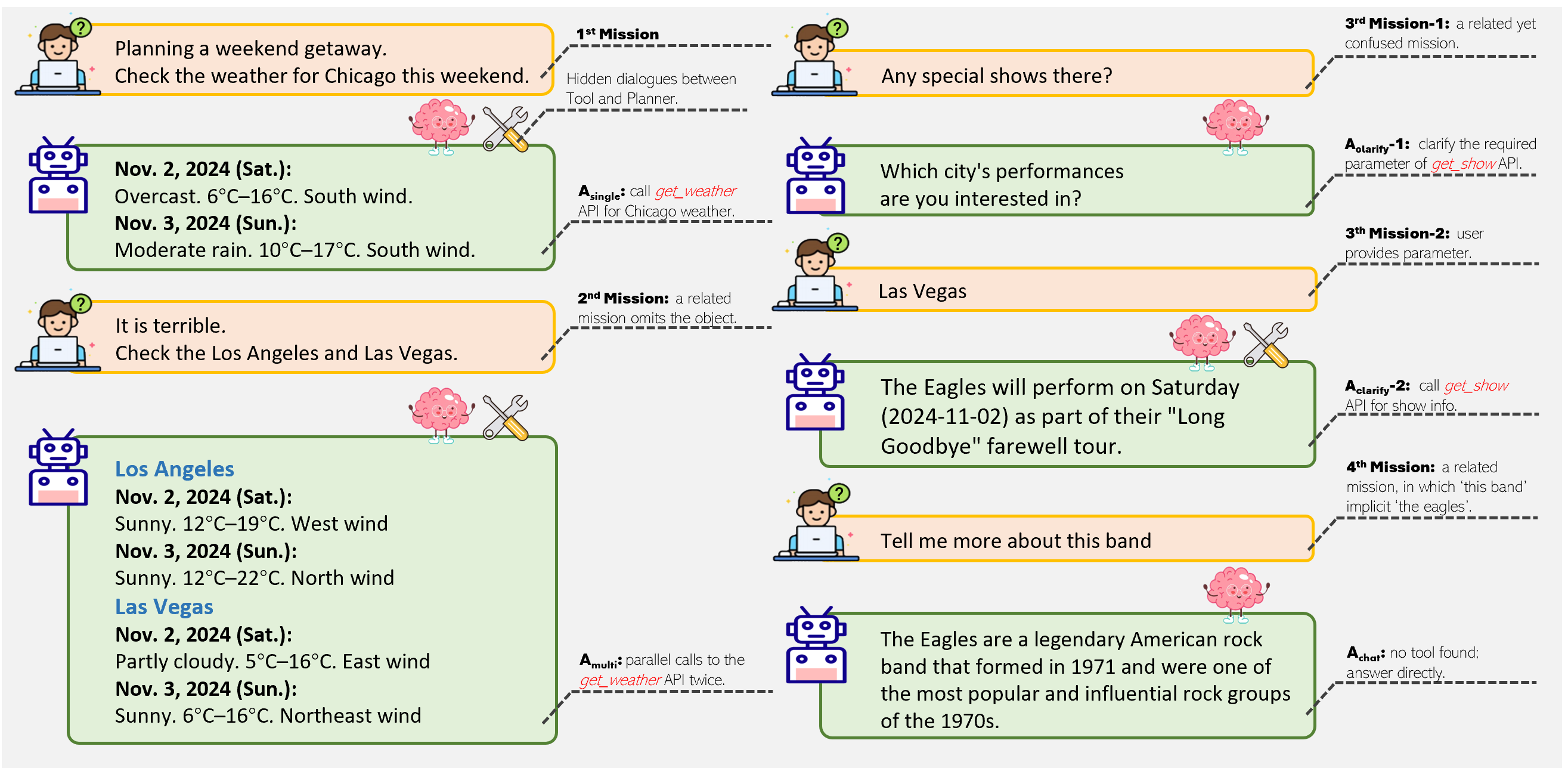} 
  \caption {A multi-task example. It contains four \textbf{related} tasks, and the task types are changing \textbf{dynamically}. This figure presents the conversation between a user and an AI. The action-observation steps are hided.}
  \label{fig:apdx_4}
\end{figure*}

\begin{figure*}[h]
  \centering
  \includegraphics[width=0.80\linewidth]{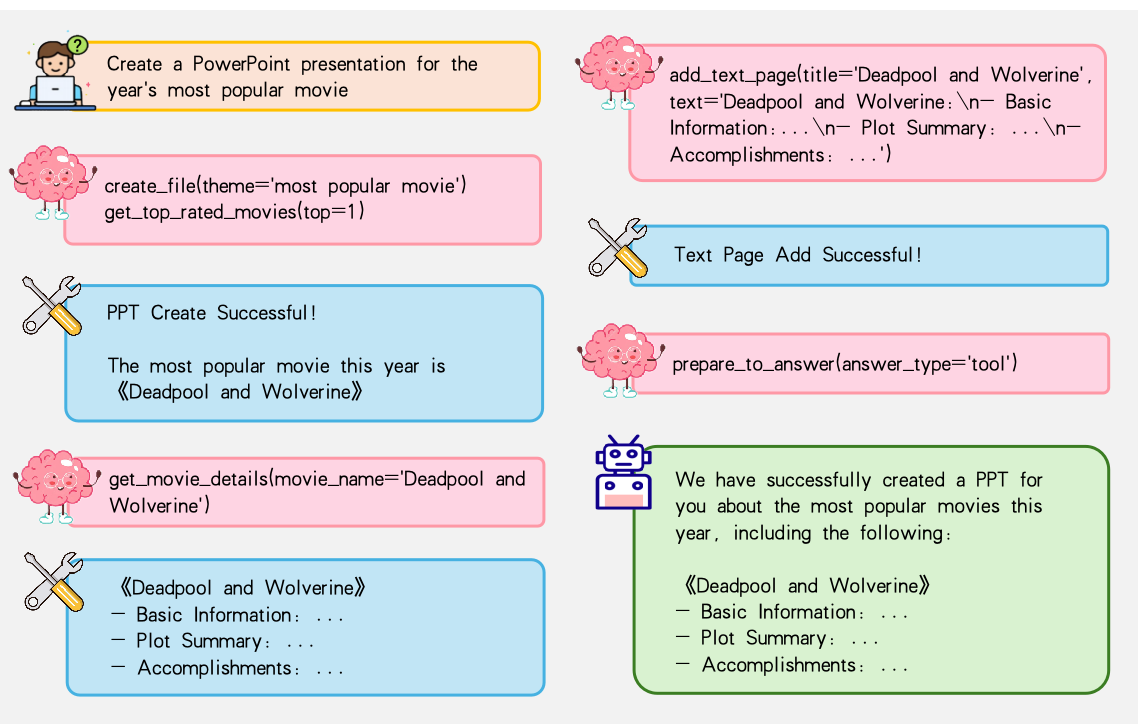} 
  \caption {An Optimal Path Example.}
  \label{fig:apdx_5}
\end{figure*}

\begin{figure*}[h]
  \centering
  \includegraphics[width=0.80\linewidth]{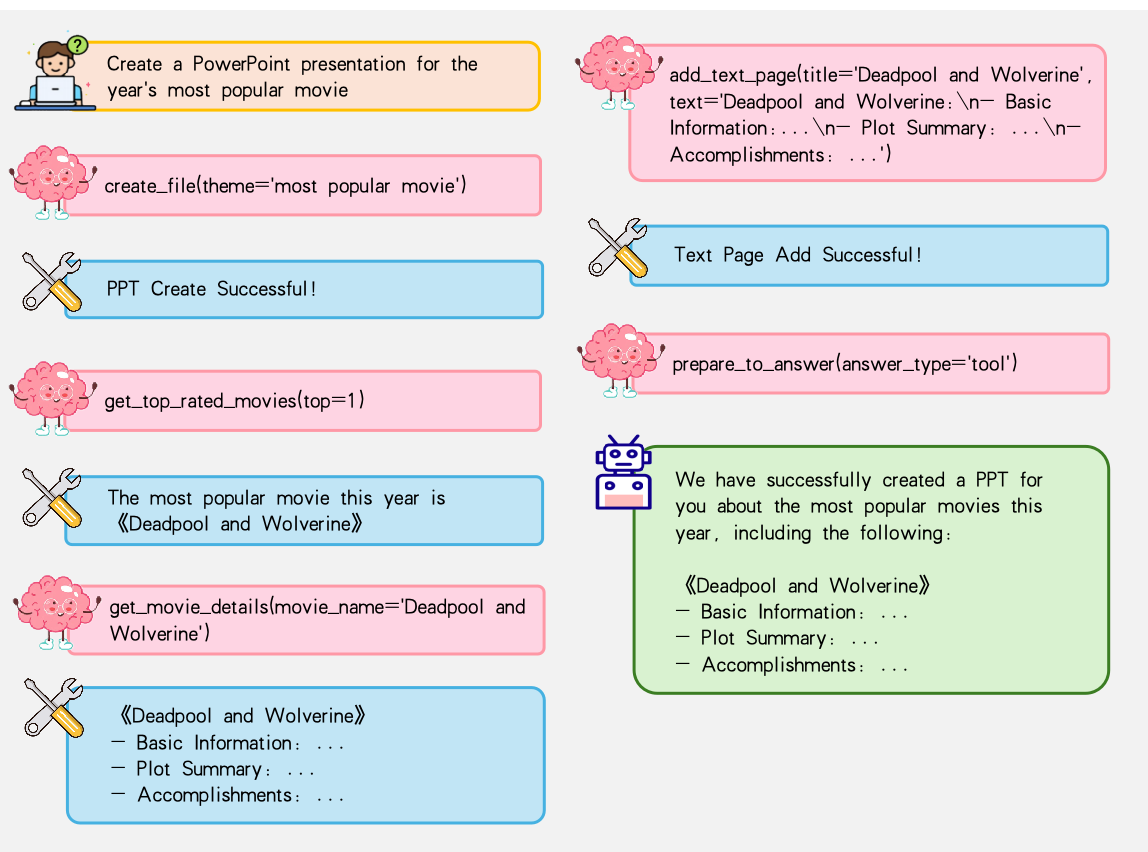} 
  \caption {A Suboptimal Path Example.}
  \label{fig:apdx_6}
\end{figure*}

To complement the examples in Section 5, we provide three additional task execution cases. Figure \ref{fig:apdx_4} illustrates a multi-task scenario with four related tasks and dynamic task-type changes, showcasing a user-AI conversation while hiding action-observation steps for clarity. Figures \ref{fig:apdx_5} and \ref{fig:apdx_6} contrast an optimal path execution with a suboptimal one, highlighting differences in efficiency and task sequencing.

\section{Part Roles Prompt of Agents}\label{appendix-c}

\subsection{Role Prompt of task Generation}
We show the role prompt of single task generation in Figure \ref{P-task}.

\subsection{Role Prompt of Planner}
We show the role prompt of Planner in Figures \ref{P-planner_1}, \ref{P-planner_2}, \ref{P-planner_3}, \ref{P-planner_4}, \ref{P-planner_5}, \ref{P-planner_6} and  \ref{P-planner_7}.

\subsection{Role Prompt of Tool}

We show the role prompt of Tool in Figures \ref{P-tool}.

\subsection{Role Prompt of AI}

We show the role prompt of AI in Figures \ref{P-AI}.

\begin{figure*}[htbp]
\begin{tcolorbox}[title={Single Tool Invocation task Generation Prompt.}]

Please act as a user interacting with a super intelligent agent.

\medskip

This super intelligent agent has access to a range of external tools and can use these tools to solve the tasks you propose.

\medskip

Next, please propose 5 tasks that you need the super intelligent agent to solve based on the

\medskip

All 5 tasks must require the use of $\{\{\{tool\}\}\}$ from the [Tool List] to be completed, and each task should only require a single call to $\{\{\{tool\}\}\}$.

\medskip

The tasks should be specific and diverse.

\medskip

Finally, please output the final result according to the [Format] without generating any extra text.

\medskip

The required parameters for tool $\{\{\{tool\}\}\}$ are: $\{\{\{tool\_required\}\}\}$, and the optional parameters are: $\{\{\{tool\_no\_required\}\}\}$.

\medskip

\medskip

[Requirements]="""

1. The description of the user's task must include information on all the required parameters needed to call $\{\{\{tool\}\}\}$. For other optional parameters, please add them as you see fit, using natural language.

2. The user's tasks should use different types of sentence structures: imperative, declarative, interrogative, etc.

3. The user's tasks should include different tones: colloquial, formal, polite, direct, etc.

4. Ensure that the length of the user's tasks varies, gradually increasing from short to long.

5. Ensure that the user's tasks involve different themes/instances, different scenarios, and different roles.

6. Extract common entities that appear in all descriptions from the [Tool List] and ensure that these entities appear in the user's tasks.

7. Do not explicitly specify the tool $\{\{\{tool\}\}\}$ in the user's tasks.

"""

\medskip

\medskip

[Tool List]="""

$\{\{\{tool\}\}\}$

"""

\medskip

\medskip

[Format]="""

\{

    \parindent=2em
    
    "task 1": "xxx",
    
    "task 2": "xxx",
    
    "task 3": "xxx",
    
    "task 4": "xxx",
    
    "task 5": "xxx",
    
    \noindent
\}

\noindent
"""

\end{tcolorbox}
\caption{Single Tool Invocation task Generation Prompt.}
\label{P-task}
\end{figure*}

\begin{figure*}[htbp]
\begin{tcolorbox}[title={Planner Decision Generation Prompt Part-1.}]

Please act as a Planner within a super intelligent agent.

\medskip

You have access to a series of external tools, and you can solve user tasks by invoking these external tools, as detailed in the [Tool List].

\medskip

You are responsible for assessing the completion status of the current user task and providing thoughts, plans, and actions to be executed.

\medskip

If the Checker\_Planner indicates `no' for correct, it means there is an issue with the decision you made in the previous round. In this case, you should regenerate your decision based on the analysis provided by the Checker\_Planner.

\medskip

However, please be mindful not to include explanations of previously generated incorrect results in your Thoughts!

\medskip

In your Plan, be sure not to mention the use of the prepare\_to\_answer tool and the ask\_user\_for\_required\_parameters tool. Instead, describe these actions in natural language, as the prepare\_to\_answer and ask\_user\_for\_required\_parameters tools are not to be exposed.

\medskip

Refer to the [Planner Output Format] for the output format.

\medskip

[Environmental Information]="""

\{\{\{env\_info\}\}\}

"""

\end{tcolorbox}
\caption{Planner Decision Generation Prompt Part-1.}
\label{P-planner_1}

\end{figure*}

\begin{figure*}[htbp]

\begin{tcolorbox}[title={Planner Decision Generation Prompt Part-2.}]

[Planner Output Format]="""

Planner:

\{

    \parindent=2em
    
    "task\_Finish": "Whether the user task is completed, fill in `yes' if completed, `no' if not completed",
    
    "Thought": "Based on the [Requirements] and [Environmental Information], follow the steps below to give the internal thought process when solving the user task. You must provide an analysis of the required and optional parameters for each tool that needs to be called.
    
First step, decompose the task, first analyze whether a tool needs to be called to complete it, and whether there is a suitable tool in the [Tool List].

If a tool needs to be called, which tool(s) should be used to complete the user task, whether one or multiple tools should be called.

If multiple tools are involved, please provide an analysis of the serial and parallel nature of multiple tools.

Second step, provide an analysis of the required and optional parameters for the first tool that needs to be called (now), in the following order.

1. First, list the required and optional parameters for each tool that needs to be called.

2. Based on the context and user task, analyze the required parameters, check which information for each tool's required parameters is provided, and explain which are provided and which are missing to ask the user.

3. Finally, analyze the optional parameters. If the user has provided information for optional parameters, briefly explain the situation; otherwise, there is no need to explain.

Note:

1. The analysis process should not be too lengthy; it needs to be concise and clear.

2. Do not have too much redundant content that is repetitive of the Plan.",

    "Plan": "Based on the [Requirements], [Environmental Information], Thought, context, and user task, provide a planning scheme.

Note:

1. When involving multiple tool calls, provide the overall plan and the plan for the first action during the first Plan, and provide the plan for the current step in subsequent dialogues.

2. The Plan is a general explanation of the Thought. The Plan does not need to set the values of the tool parameters; it only needs to explain which tools should be called to complete what tasks, only the purpose of calling the tools.

3. The format of the Plan needs to be consistent with the example given in the [Requirements].

4. Do not have too much redundant content that is repetitive of the Thought.",

    "Action\_List": [

        \parindent=4em
        
        \{  

            \parindent=6em
            
            "name": "Based on the [Requirements], provide the action to be taken, i.e., the selected tool name",
            
            "arguments": "Based on the [Requirements], [Environmental Information], and [Tool List], provide the input parameters for the action to be taken, i.e., the tool's input parameters. Note: 1. Optional parameters not specified by the user do not need to be provided. 2. Use the JSON format in terms of format, use a dictionary object, do not use strings, and there is no need to provide comments for the parameters",
            
            "tool\_call\_purpose": "The purpose of the tool call"
            
        \parindent=4em
        \}

    \parindent=2em
    ]

\noindent\}

\noindent"""

\end{tcolorbox}
\caption{Planner Decision Generation Prompt Part-2.}
\label{P-planner_2}
\end{figure*}

\begin{figure*}[htbp]

\begin{tcolorbox}[title={Planner Decision Generation Prompt Part-3.}]

[Requirements]="""

*** Special Attention ***

1. When making a decision, please ensure that the tool you invoke from the [Tool List] is suitable for solving the user's task based on the definition of the tools in the list. Do not force the use of inappropriate tools to solve the user's task; instead, call the appropriate tool from the [Tool List] according to the user's task.

\medskip

2. Ensure that the Action\_List you provide does not contradict the Plan you have set out. The order of tools in the given Action\_List should be consistent with the sequence planned in the Plan.

\medskip

3. For optional parameters, you only need to fill them in if the user has provided a value that is different from the default or if there is no default value. Otherwise, there is no need to include them in the arguments.

\medskip

*** The prepare\_to\_answer tool needs to be called in the following two scenarios: ***

1. If you believe that the user's task can be completed, then call the prepare\_to\_answer tool to provide a summary response, with the answer\_type parameter set to `tool'.

\medskip

2. If you believe that the user's task does not require the use of any tools from the [Tool List] or that there is no suitable tool to solve the user's task and it can be answered directly, then call the prepare\_to\_answer tool, with the answer\_type parameter set to `chat'.

\medskip

Note:

1) The absence of a suitable tool in the [Tool List] to solve the user's task does not mean that you lack the ability to answer. Please respond based on the context information and the knowledge you possess. Do not excessively refuse to answer, nor imagine knowledge you do not have. Only refuse to answer when you cannot respond based on the context information and your own knowledge.

\medskip

2) The absence of a suitable tool in the [Tool List] to solve the user's task also includes the following situation:

\medskip

First, analyze the common entities that appear in each tool. For example, some tools can only query data related to a certain entity A. If the user asks about entity B, it also means that there is no suitable tool.

\medskip

For instance:

- If the tools in the [Tool List] can only query and analyze population data for Denmark, and the user asks for population data for Sweden, then you should also call the prepare\_to\_answer tool.

\medskip

- If the tools in the [Tool List] can only query weather data for China, including current and historical weather, and the user asks for weather data for the United States, then you should also call the prepare\_to\_answer tool.

\end{tcolorbox}
\caption{Planner Decision Generation Prompt Part-3.}
\label{P-planner_3}
\end{figure*}

\begin{figure*}[htbp]

\begin{tcolorbox}[title={Planner Decision Generation Prompt Part-4.}]

*** There are four scenarios in which the ask\_user\_for\_required\_parameters tool needs to be invoked: ***

1. If you believe that a user's task requires the use of a tool from the [Tool List], but the user's task is missing some required parameters from the tool, and the user needs to provide the necessary information, then invoke the ask\_user\_for\_required\_parameters tool. Please do not hallucinate parameters.

\medskip

2. Please note that you are unable to deduce the values of some tool parameters on your own and will need to invoke the ask\_user\_for\_required\_parameters tool to ask the user. Please do not hallucinate parameters.

\medskip

For example:

1) For the timestamp parameter, you do not have the ability to deduce the timestamp based on time. However, you can deduce other time-related parameters (start\_time, end\_time, etc.) on your own based on [Environmental Information], without needing to invoke the ask\_user\_for\_required\_parameters tool.

2) For ID-type parameters (station\_id, product\_id, etc.), you do not have the ability to deduce the corresponding ID based on the name.

\medskip

3. Based on the context of the conversation, if you have already asked the user once to provide the necessary information but the user still has not provided all the required parameters, then please continue to invoke the ask\_user\_for\_required\_parameters tool.

\medskip

4. If the user provides incomplete parameter values, such as the tool parameter being an IP address (ip\_address), but the user provides an incomplete IP address (e.g., 192.22), please continue to use the ask\_user\_for\_required\_parameters tool to ask the user for the complete IP address.

\medskip

Finally, if you confirm the need to invoke the ask\_user\_for\_required\_parameters tool, provide the inquiry plan in the format: "Ask the user to provide xxx, in order to invoke the xxx tool to xxx" in the Plan.

\end{tcolorbox}
\caption{Planner Decision Generation Prompt Part-4.}
\label{P-planner_4}
\end{figure*}

\begin{figure*}[htbp]

\begin{tcolorbox}[title={Planner Decision Generation Prompt Part-5.}]

*** There are eight scenarios in which multiple tools need to be invoked: ***

If a user task involves invoking multiple tools, please first analyze the dependency relationships between the multiple invocation tools. For tools that do not have invocation dependencies, perform concurrent invocations, and for tools that do have invocation dependencies, perform serial invocations.
Specifically, you can handle each of the following eight scenarios separately:

\medskip

Concurrent invocation scenarios:

\medskip

1. If you determine that the user task requires multiple invocations of the same tool A, but with different parameters for each invocation of tool A, then please invoke tool A concurrently and provide the concurrent invocation plan in the Plan in the format: "Concurrently invoke tool A N times for xxx."

\medskip

2. If you determine that the user task requires the invocation of different tools, such as tools A and B, and there is no dependency between tool A and B, then please invoke tools A and B concurrently, and provide the concurrent invocation plan in the Plan in the format: "Concurrently invoke tool A for xxx, while invoking tool B for xxx."

\medskip

Serial invocation scenarios:

\medskip

3. If you determine that the user task requires the invocation of different tools, such as tools A, B, and C, and there are dependencies between these tools, then please invoke tools A, B, and C serially, and provide the serial invocation plan in the Plan in the format: "First, invoke tool A for xxx. Then, invoke tool B for xxx. Next, invoke tool C for xxx. Now, I will invoke tool A for xxx."

\medskip

Serial invocation has the following two dependency scenarios:

\medskip

3.1. Parameter dependency: For example, before invoking tool C, it is necessary to first invoke tool B to obtain the result as an input parameter, and before invoking tool B, it is necessary to first invoke tool A to obtain the result as an input parameter. Therefore, you need to first complete the invocation of tool A to obtain the result, use it as the input parameter for invoking tool B, and after obtaining the result from tool B, use it as the input parameter for invoking tool C, i.e., please invoke tools A, B, and C serially.

\medskip

3.2. Logical dependency: Even if there is no parameter dependency between the invocation of tools A, B, and C, but there is a logical dependency, such as logically needing to invoke tool B before tool C, and tool A before tool B, then please also invoke tools A, B, and C serially.

\end{tcolorbox}
\caption{Planner Decision Prompt Generation Part-5.}
\label{P-planner_5}
\end{figure*}

\begin{figure*}[htbp]

\begin{tcolorbox}[title={Planner Decision Generation Prompt Part-6.}]

Combined serial and concurrent invocation scenarios:

\medskip

4. If you determine that the user task requires the invocation of different tools, such as tools A, B, and C, and tool C depends on the invocation of tools A and B, but there is no dependency between tools A and B, then please invoke tools A and B concurrently, followed by the serial invocation of tool C, and provide the combined serial and concurrent invocation plan in the Plan in the format: "Concurrently invoke tools A and B for xxx and xxx, respectively. Then, invoke tool C for xxx. Now, I will concurrently invoke tools A and B for xxx and xxx."

\medskip

5. If you determine that the user task requires the invocation of different tools, such as tools A, B, and C, and tools B and C depend on the invocation of tool A, but there is no dependency between tools B and C, then please first invoke tool A serially, followed by the concurrent invocation of tools B and C, and provide the combined serial and concurrent invocation plan in the Plan in the format: "First, invoke tool A for xxx. Then, concurrently invoke tools B and C for xxx and xxx, respectively. Now, I will invoke tool A for xxx."

\medskip

6. If you determine that the user task requires the invocation of different tools, such as tools A and B, and there is a dependency between tools A and B, and tool A needs to be invoked multiple times, then please first invoke tool A concurrently multiple times, followed by the serial invocation of tool B, and provide the combined serial and concurrent invocation plan in the Plan in the format: "First, concurrently invoke tool A N times for xxx. Then, invoke tool B for xxx. Now, I will concurrently invoke tool A N times for xxx."

\medskip

7. If you determine that the user task requires the invocation of different tools, such as tools A and B, and there is a dependency between tools A and B, and tool B needs to be invoked multiple times, then please first invoke tool A serially, followed by the concurrent invocation of tool B multiple times, and provide the combined serial and concurrent invocation plan in the Plan in the format: "First, invoke tool A for xxx. Then, concurrently invoke tool B N times for xxx. Now, I will invoke tool A for xxx."

\medskip

Special scenarios:

\medskip

8. The tools prepare\_to\_answer and ask\_user\_for\_required\_parameters cannot be invoked concurrently with other tools and need to be invoked serially.

\end{tcolorbox}
\caption{Planner Decision Generation Prompt Part-6.}
\label{P-planner_6}
\end{figure*}

\begin{figure*}[htbp]

\begin{tcolorbox}[title={Planner Decision Generation Prompt Part-7.}]

Please also note:

\medskip

1. The dependency relationship between tool invocations refers to the necessity of completing the call to Tool A before running the call to Tool B.

\medskip

2. For multiple invocations of the same tool, it is necessary to carefully analyze the dependency relationship of each call, noting that even two calls to the same tool may be interdependent.

\medskip

3. If you state in your Thought and Plan that tools need to be called in sequence, then the number of tools to be called in your given Action\_List cannot exceed one, otherwise, there will be a logical contradiction!

\medskip

4. If you cannot ensure that parallel calls to multiple tools A, B, C will not have parameter dependencies and logical dependencies, then please call multiple tools A, B, C in sequence!

\medskip

*** Special Circumstances ***

\medskip

In the following three cases, there is no need to call the ask\_user\_for\_required\_parameters tool:

\medskip

1. If a tool's parameter is a country's ISO code, and the user's task mentions a specific country, such as China, you can directly deduce China's ISO code and fill it in.

\medskip

2. If a tool's parameter is a longitude or latitude value, and the user's task mentions a specific location, such as Beijing, you can directly deduce the approximate longitude and latitude values for Beijing and fill them in.

\medskip

3. If a tool's parameter is a time-related parameter (such as start\_time, end\_time, or other parameters that include year, month, and day) and not a timestamp type, you can deduce it based on the current time in the [Environmental Information] and fill it in. At the same time, you need to explain in your Thought how you deduced the value of the time-related parameter based on the current time.

\medskip

*** Other Notes: ***

\medskip

1. Be sure not to provide comments for parameters, as providing parameter comments will cause JSON to fail to parse.

"""

\medskip

\{\{\{all\_tool\_required\_info\}\}\}

\medskip

[Tool List]="""

\{\{\{tools\}\}\}

"""

\end{tcolorbox}
\caption{Planner Decision Generation Prompt Part-7.}
\label{P-planner_7}
\end{figure*}

\begin{figure*}[htbp]

\begin{tcolorbox}[title={Tool Feedback Generation Prompt.}]

Please act as an external tool, Tool, within a super intelligent agent. These external tools can be used to solve user tasks, as detailed in the [Tool List].

\medskip

Based on the tool name and input parameters output by the super intelligent agent's Planner, simulate the execution results of the tool.

\medskip

If there are multiple tools in the Action\_List provided by the Planner, please simulate each one separately, ensuring the number matches the Action\_List, and store the results in the Observation\_List.
Refer to the [Tool Output Format] for the output format.

\medskip

[Environmental Information]="""

\{\{\{env\_info\}\}\}

"""

\medskip

\medskip

[Tool Invocation Result Requirements]="""

1. Validate the HTTP method and parameters in the request according to the OpenAPI specification.

2. Generate a response that strictly follows the format specified in the OpenAPI specification and ensure it is in JSON format.

3. The response should contain real data, avoiding the use of placeholders.

4. Handle edge cases by providing appropriate error responses.

5. For requests without length limitations, such as the GET method, ensure the response returns 3 to 5 samples, and be careful not to use ellipses like // xxx, ... to omit sample information, as it must conform to JSON format, otherwise it will cause JSON parsing errors!

6. Try to simulate responses in English.

"""

\medskip

[Tool List]="""

\{\{\{tools\}\}\}

"""

\medskip

[Tool Output Format]="""

Tool:

\{

    \parindent=2em

    "Observation\_List": [

        \parindent=4em
        
        \{

            \parindent=6em
        
            "status\_code": "Refer to [Tool Invocation Result Requirements] for the HTTP response status code",
            
            "response": "Refer to [Tool Invocation Result Requirements] to simulate the result of the action execution. Ensure your response is in JSON format, contains real data, and complies with the OpenAPI specification format."

        \parindent=4em    
        \}

    \parindent=2em
    ]
 
\noindent
\}

\noindent
"""

\end{tcolorbox}
\caption{Tool Feedback Generation Prompt.}
\label{P-tool}
\end{figure*}

\begin{figure*}[htbp]
\begin{tcolorbox}[title={AI Feedback Generation Prompt.}]

Please act as an Agent assistant within a super intelligent agent, which has a series of external tools. The Planner within the super intelligent agent can solve user tasks by calling external tools, as detailed in the [Tool List].

\medskip

You are responsible for interacting with the user. Based on the results returned by the Planner and Tool, combined with the user task and the context of the conversation, you provide answers, and only your answers will be displayed to the user.

\medskip

Refer to the [Agent Assistant Output Format] for the output format.

\medskip

[Environmental Information]="""

\{\{\{env\_info\}\}\}

"""

\medskip

[Agent Assistant Output Format]="""

Agent Assistant: According to the [Requirements], reply to the most recent round of content starting with "User:" in the context conversation information (do not repeat this sentence).

"""

\medskip

[Requirements]="""

1. The reply must start with "Agent Assistant:".

2. Summarize the user task from the most recent round starting with "User:" based on the context conversation information.

3. Use markdown format, and be sure to pay attention to the layout to make it look neat, with two line breaks between paragraphs.

4. Pay special attention! If the Observation given by the Tool is a list, and each item in the list has its own ID, such as xxx\_id or xxxId, then when summarizing the reply, please retain these IDs for each item and inform the user!

5. Reply in English.

"""

\medskip

\{\{\{all\_tool\_required\_info\}\}\}

\medskip

[Tool List]="""

\{\{\{tools\}\}\}

"""

\end{tcolorbox}
\caption{AI Feedback Generation Prompt.}
\label{P-AI}

\end{figure*}


\section{Demo Cases}
\subsection{Policy Demo}

\subsubsection{Four Policies}\label{demo-1}
We first show the demo of 4 policies, namely $P_{single}, P_{multi}, P_{clarify}, P_{chat}$
in Tables \ref{demo-1-1}, \ref{demo-1-2}, \ref{demo-1-3}, \ref{demo-1-4}.

\begin{table*}[ht]
\centering
\caption{Demo of policy $P_{single}$. Data ID: 2c6ee79d-168d-45cf-a0d8-459243976d73}
\label{demo-1-1}
\begin{tabular}{|c|p{12cm}|} 
\hline
\textbf{Role} & \textbf{Dialogue} \\
\hline
\multirow{7}{*}{\textbf{User}} & 
  \medskip
  I want to travel to other places this weekend. \\
  & Please help me check the weather in Chicago \\
  & for the two days of the weekend.  
  \medskip
\\ \hline
\multirow{7}{*}{ \textbf{Action}} & 
  \medskip
    [ \\
    & \quad \{ \\
    & \quad \quad "name": "getCityForecast", \\
    & \quad \quad "arguments": \{ \\
    & \quad \quad \quad "city": "Chicago", \\
    & \quad \quad \quad "startDate": "2024-07-13", \\
    & \quad \quad \quad "endDate": "2024-07-14" \\
    & \quad \quad \} \\
    & \quad \} \\
    & ]
  \medskip
\\ \hline
\multirow{15}{*}{\textbf{Observation}} &  
  \medskip
    [ \\
    & \quad \{ \\
    & \quad \quad "response": \{ \\
    & \quad \quad \quad "forecast": [ \\
    & \quad \quad \quad \quad \{ \\
    & \quad \quad \quad \quad \quad "date": "2024-07-13", \\
    & \quad \quad \quad \quad \quad "humidity": "80\%", \\
    & \quad \quad \quad \quad \quad "temperature": \{ \\
    & \quad \quad \quad \quad \quad \quad "high": "25°C", \\
    & \quad \quad \quad \quad \quad \quad "low": "18°C" \\
    & \quad \quad \quad \quad \quad \}, \\
    & \quad \quad \quad \quad \quad "weather": "Light rain" \\
    & \quad \quad \quad \quad \}, \\
    & \quad \quad \quad \quad \{ \\
    & \quad \quad \quad \quad \quad "date": "2024-07-14", \\
    & \quad \quad \quad \quad \quad "humidity": "85\%", \\
    & \quad \quad \quad \quad \quad "temperature": \{ \\
    & \quad \quad \quad \quad \quad \quad "high": "23°C", \\
    & \quad \quad \quad \quad \quad \quad "low": "17°C" \\
    & \quad \quad \quad \quad \quad \}, \\
    & \quad \quad \quad \quad \quad "weather": "Moderate rain" \\
    & \quad \quad \quad \quad \} \\
    & \quad \quad \quad ], \\
    & \quad \quad \quad "status\_code": 200 \\
    & \quad \quad \} \\
    & \quad \} \\
    & ]
  \medskip
\\ \hline
\multirow{5}{*}{\textbf{Summary}} & 
  \medskip
  Here is the weather forecast for Chicago this weekend:  
  \begin{itemize} 
    \item On Saturday, July 13th, the weather will be light rain with a high of 25°C and a low of 18°C. The humidity will be 80\%.  
    \item On Sunday, July 14th, the weather will be moderate rain with a high of 23°C and a low of 17°C. The humidity will be 85\%.  
  \end{itemize}  
  \medskip
\\ \hline
\end{tabular}
\end{table*}

\begin{table*}[ht]
\centering
\small
\caption{Demo of policy $P_{multi}$. Data ID: 967a29a4-73a6-47ca-880f-8135992f1333
}
\label{demo-1-2}
\begin{tabular}{|c|p{12cm}|} 
\hline
\textbf{Role} & \textbf{Dialogue} \\
\hline
\multirow{6}{*}{\textbf{User}} & 
  \medskip
  Please help me find examples of the word 'ubiquitous' used in technology articles and tell me its synonyms.  
  \medskip
\\ \hline
\multirow{7}{*}{ \textbf{Action}} & 
  \medskip
  [ \\
  & \quad \{ \\
  & \quad \quad "name": "getWordExamples", \\
  & \quad \quad "arguments": \{ \\
  & \quad \quad \quad "word": "ubiquitous", \\
  & \quad \quad \quad "useCase": "technology articles" \\
  & \quad \quad \} \\
  & \quad \}, \\
  & \quad \{ \\
  & \quad \quad "name": "getWordSynonyms", \\
  & \quad \quad "arguments": \{ \\
  & \quad \quad \quad "word": "ubiquitous" \\
  & \quad \quad \} \\
  & \quad \} \\
  & ]
  \medskip
\\ \hline
\multirow{18}{*}{\textbf{Observation}} &  
  \medskip
  [ \\
  & \quad \{ \\
  & \quad \quad "response": \{ \\
  & \quad \quad \quad "examples": [ \\
  & \quad \quad \quad \quad \{ \\
  & \quad \quad \quad \quad \quad "sentence": "The internet has become a ubiquitous part of everyday life, providing access to information and communication no matter where you are." \\
  & \quad \quad \quad \quad \}, \\
  & \quad \quad \quad \quad \{ \\
  & \quad \quad \quad \quad \quad "sentence": "With the advent of smartphones, technology has become ubiquitous, enabling constant connectivity and digital interaction." \\
  & \quad \quad \quad \quad \}, \\
  & \quad \quad \quad \quad \{ \\
  & \quad \quad \quad \quad \quad "sentence": "Cloud computing has made data storage and processing power ubiquitous, allowing businesses of all sizes to harness advanced technology." \\
  & \quad \quad \quad \quad \} \\
  & \quad \quad \quad ], \\
  & \quad \quad \quad "status\_code": 200 \\
  & \quad \quad \}, \\
  & \quad \quad \{ \\
  & \quad \quad \quad "response": \{ \\
  & \quad \quad \quad \quad "synonyms": [ \\
  & \quad \quad \quad \quad \quad "omnipresent", \\
  & \quad \quad \quad \quad \quad "everywhere", \\
  & \quad \quad \quad \quad \quad "pervasive", \\
  & \quad \quad \quad \quad \quad "universal", \\
  & \quad \quad \quad \quad \quad "widespread" \\
  & \quad \quad \quad \quad ], \\
  & \quad \quad \quad \quad "status\_code": 200 \\
  & \quad \quad \quad \} \\
  & \quad \quad \} \\
  & \quad \} \\
  & ]
  \medskip
\\ \hline
\multirow{6}{*}{\textbf{Summary}} & 
  \medskip
  Here are examples of 'ubiquitous' in technology articles:  
  \begin{itemize} 
    \item The internet has become a ubiquitous part of everyday life, providing access to information and communication no matter where you are.  
    \item With the advent of smartphones, technology has become ubiquitous, enabling constant connectivity and digital interaction.  
    \item Cloud computing has made data storage and processing power ubiquitous, allowing businesses of all sizes to harness advanced technology.  
  \end{itemize}  

  Here are synonyms for 'ubiquitous': omnipresent, everywhere, pervasive, universal, widespread
  \medskip
\\ \hline
\end{tabular}
\end{table*}

\begin{table*}[ht]
\centering
\caption{Demo of policy $P_{chat}$. Data ID: 05fe357d-27ff-4f04-b46a-309eb8dc5b5e}
\label{demo-1-3}
\begin{tabular}{|c|p{12cm}|} 
\hline
\textbf{Role} & \textbf{Dialogue} \\
\hline
\multirow{5}{*}{\textbf{User}} & 
  \medskip
  When designing API interfaces, how can I ensure a balance between security and efficiency?  
  \medskip
\\ \hline
\multirow{40}{*}{\textbf{Summary}} & 
  \medskip
  Balancing security and efficiency in API design is crucial. Here are some key considerations:  
  \begin{enumerate} 
    \item \textbf{Authentication and Authorization:}  
      Implement strong authentication mechanisms (like OAuth 2.0, JWTs) to verify the identity of the caller. Use granular authorization to ensure users only access resources they are permitted to. Be efficient by caching tokens or using stateless tokens where appropriate.  
    \item \textbf{Input Validation:}  
      Validate all incoming data rigorously to prevent injection attacks (SQL, XSS, etc.). While validation adds a small overhead, it's essential for security and prevents processing invalid or malicious data.  
    \item \textbf{Rate Limiting and Throttling:}  
      Protect your API from abuse and Denial-of-Service (DoS) attacks by implementing rate limiting. This restricts the number of requests a client can make within a certain time frame. This enhances security by preventing brute-force attacks and improves efficiency by preventing server overload.  
    \item \textbf{Use HTTPS:}  
      Always use HTTPS to encrypt data in transit, protecting sensitive information from eavesdropping. The performance overhead of TLS is generally acceptable for most applications today due to hardware acceleration.  
    \item \textbf{Minimize Data Transfer:}  
      Only return necessary data in API responses. Over-fetching data increases bandwidth usage and processing time for both the server and the client. Allow clients to specify fields if possible (e.g., using GraphQL or sparse fieldsets in REST).  
    \item \textbf{Caching:}  
      Implement caching strategies at various levels (client-side, server-side, API gateway) to reduce the load on your backend services and speed up response times for frequently requested data. Ensure cached data respects security constraints.  
    \item \textbf{API Gateway:}  
      Use an API Gateway to centralize concerns like authentication, rate limiting, logging, and monitoring. This offloads these tasks from your backend services, making them simpler and more efficient, while also providing a central point for enforcing security policies.  
    \item \textbf{Error Handling:}  
      Implement informative but not overly detailed error messages. Avoid revealing sensitive information in error responses.  
    \item \textbf{Logging and Monitoring:}  
      Implement comprehensive logging and monitoring to detect suspicious activity, performance bottlenecks, and security incidents.  
  \end{enumerate}  

  In essence, security should be built-in from the start, not added as an afterthought. Many security measures, like rate limiting and caching, also directly contribute to better performance and efficiency. It's about finding the right balance of controls based on the sensitivity of the data and the expected traffic volume.  
  \medskip
\\ \hline
\end{tabular}
\end{table*}

\begin{table*}[ht]
\centering
\caption{Demo of policy $P_{clarify}$. Data ID: f679d978-de87-4a56-820b-28dfae7c18b5}
\label{demo-1-4}
\begin{tabular}{|c|p{12cm}|} 
\hline
\textbf{Role} & \textbf{Dialogue} \\
\hline
\multirow{7}{*}{\textbf{User}} & 
  \medskip
  Please mark the locations of the library and the city hall on the map. \\
  & The library is located at latitude 34.052235 and longitude -118.243683. \\
  & The city hall is located at latitude 34.053690 and longitude -118.242766. 
  \medskip
\\ \hline
\multirow{6}{*}{\textbf{Action}} & 
  \medskip
  Could you please provide the ID of the map where you would like to mark the locations?  
  \medskip
\\ \hline
\multirow{4}{*}{\textbf{User}} & 
  \medskip
  Use the map with map number "map\_12345".
  \medskip
\\ \hline
\multirow{12}{*}{ \textbf{Action}} & 
  \medskip
    [ \\  
    & \quad \{ \\  
    & \quad \quad "name": "addCustomMarker", \\  
    & \quad \quad "arguments": \{ \\  
    & \quad \quad \quad "mapId": "map\_12345", \\  
    & \quad \quad \quad "position": \{ \\  
    & \quad \quad \quad \quad "latitude": 34.052235, \\  
    & \quad \quad \quad \quad "longitude": -118.243683 \\  
    & \quad \quad \quad \}, \\  
    & \quad \quad \quad "title": "Library" \\  
    & \quad \quad \} \\  
    & \quad \}, \\ 
    & \quad \{ \\  
    & \quad \quad "name": "addCustomMarker", \\  
    & \quad \quad "arguments": \{ \\  
    & \quad \quad \quad "mapId": "map\_12345", \\  
    & \quad \quad \quad "position": \{ \\  
    & \quad \quad \quad \quad "latitude": 34.05369, \\  
    & \quad \quad \quad \quad "longitude": -118.242766 \\  
    & \quad \quad \quad \}, \\  
    & \quad \quad \quad "title": "City Hall" \\  
    & \quad \quad \} \\  
    & \quad \} \\  
    & ]
  \medskip
\\ \hline
\multirow{18}{*}{\textbf{Observation}} &  
  \medskip
    [ \\  
    & \quad \{ \\  
    & \quad \quad "response": \{ \\  
    & \quad \quad \quad "mapId": "map\_12345", \\  
    & \quad \quad \quad "markerId": "marker\_1", \\  
    & \quad \quad \quad "position": \{ \\  
    & \quad \quad \quad \quad "latitude": 34.052235, \\  
    & \quad \quad \quad \quad "longitude": -118.243683 \\  
    & \quad \quad \quad \}, \\  
    & \quad \quad \quad "title": "Library" \\  
    & \quad \quad \}, \\  
    & \quad \quad "status\_code": 200 \\  
    & \quad \}, \\
    & \quad \{ \\  
    & \quad \quad "response": \{ \\  
    & \quad \quad \quad "mapId": "map\_12345", \\  
    & \quad \quad \quad "markerId": "marker\_2", \\  
    & \quad \quad \quad "position": \{ \\  
    & \quad \quad \quad \quad "latitude": 34.05369, \\  
    & \quad \quad \quad \quad "longitude": -118.242766 \\  
    & \quad \quad \quad \}, \\  
    & \quad \quad \quad "title": "City Hall" \\  
    & \quad \quad \}, \\  
    & \quad \quad "status\_code": 200 \\  
    & \quad \} \\  
    & ]
  \medskip
\\ \hline
\multirow{5}{*}{\textbf{Summary}} & 
  \medskip
  The library and the city hall have been marked on the map with ID "map\_12345".  
  \medskip
\\ \hline
\end{tabular}
\end{table*}

\subsubsection{Three Type of $P_{multi}$}\label{demo-2}
Here, we show the demo of 3 types of$P_{multi}$ , namely $P_{multi}^{S}, P_{multi}^{P}, P_{multi}^{S+P}$
in Tables \ref{demo-2-1}, \ref{demo-2-2}, \ref{demo-2-3}.

\begin{table*}[ht]
\centering
\caption{Demo of policy $P_{multi}^{S}$. Data ID: e619b53b-8402-4f43-bc48-250f0c58c2dc
}
\small
\label{demo-2-1}
\begin{tabular}{|c|p{12cm}|} 
\hline
\textbf{Role} & \textbf{Dialogue} \\
\hline
\multirow{7}{*}{\textbf{User}} & 
  \medskip
  I want to know more about the detailed information and historical background of Nyungwe Forest National Park in Rwanda.
  \medskip
\\ \hline
\multirow{8}{*}{ \textbf{Action}} & 
  \medskip
  [ \\
  & \quad \{ \\
  & \quad \quad "name": "searchLocation", \\
  & \quad \quad "arguments": \{ \\
  & \quad \quad \quad "query": "Nyungwe Forest National Park" \\
  & \quad \quad \} \\
  & \quad\} \\
  &]
  \medskip
\\ \hline
\multirow{11}{*}{\textbf{Observation}} &  
    \medskip
    [ \\  
    & \quad \{ \\  
    & \quad \quad "response": \{ \\  
    & \quad \quad \quad "locations": [ \\  
    & \quad \quad \quad \quad \{ \\  
    & \quad \quad \quad \quad \quad "coordinates": \{ \\  
    & \quad \quad \quad \quad \quad \quad "latitude": -2.4808, \\  
    & \quad \quad \quad \quad \quad \quad "longitude": 29.2 \\  
    & \quad \quad \quad \quad \quad \}, \\  
    & \quad \quad \quad \quad \quad "description": "Nyungwe Forest National Park is a national park in Rwanda, known for its rich biodiversity and virgin forests.", \\  
    & \quad \quad \quad \quad \quad "id": 2, \\  
    & \quad \quad \quad \quad \quad "name": "Nyungwe Forest National Park", \\  
    & \quad \quad \quad \quad \quad "type": "national\_park" \\  
    & \quad \quad \quad \quad \} \\  
    & \quad \quad \quad ] \\  
    & \quad \quad \} \\  
    & \quad \} \\  
    & ]
    \medskip
\\ \hline
\multirow{8}{*}{ \textbf{Action}} & 
    \medskip
    [ \\  
    & \quad \{ \\  
    & \quad \quad "name": "getLocationDetails", \\  
    & \quad \quad "arguments": \{ \\  
    & \quad \quad \quad "locationId": 2 \\  
    & \quad \quad \} \\  
    & \quad \} \\  
    & ]
    \medskip
\\ \hline
\multirow{39}{*}{\textbf{Observation}} &  
    \medskip
    [ \\
    & \quad \{ \\
      & \quad \quad "response": \{ \\
      & \quad \quad \quad "location": \{ \\
      & \quad \quad \quad \quad "address": "Nyamagabe, Rwanda", \\
      & \quad \quad \quad \quad "contact": \{ \\
      & \quad \quad \quad \quad \quad "email": "info@nyungwepark.rw", \\
      & \quad \quad \quad \quad \quad "phone": "+250 788 123 789" \\
      & \quad \quad \quad \quad \}, \\
      & \quad \quad \quad \quad "coordinates": \{ \\
      & \quad \quad \quad \quad \quad "latitude": -2.4808, \\
      & \quad \quad \quad \quad \quad "longitude": 29.2 \\
      & \quad \quad \quad \quad \}, \\
      & \quad \quad \quad \quad "description": "Nyungwe Forest National Park is a national park in Rwanda, known for its rich biodiversity and virgin forests. There are more than 1,000 plant species, 300 bird species and 75 mammal species in the park.", \\
      & \quad \quad \quad \quad "history": "Nyungwe Forest National Park was established in 2004 and is one of the largest national parks in Rwanda. The history of the park can be traced back millions of years and it is one of the oldest rainforests in Africa. ", \\
      & \quad \quad \quad \quad "id": 2, \\
      & \quad \quad \quad \quad "name": "Nyungwe Forest National Park", \\
      & \quad \quad \quad \quad "province": "Province de l'Ouest", \\
      & \quad \quad \quad \quad "provinceId": 3, \\
      & \quad \quad \quad \quad "type": "national\_park" \\
      & \quad \quad \quad \} \\
      & \quad \quad \} \\
      & \quad \} \\ 
    & ]
    \medskip
\\ \hline
\multirow{7}{*}{\textbf{Summary}} & 
  \medskip
  Here is the detailed information and historical background about Nyungwe Forest National Park:  

  \textbf{Name:} Nyungwe Forest National Park  
  \textbf{Description:}  
  Nyungwe Forest National Park is a national park in Rwanda, known for its rich biodiversity and virgin forests.....
  \medskip
\\ \hline
\end{tabular}
\end{table*}

\begin{table*}[ht]
\centering
\scriptsize 

\caption{Demo of policy $P_{multi}^{P}$. Data ID: e619b53b-8402-4f43-bc48-250f0c58c2dc
}
\label{demo-2-2}
\begin{tabular}{|c|p{12cm}|} 

\hline
\textbf{Role} & \textbf{Dialogue} \\
\hline
\multirow{7}{*}{\textbf{User}} & 
  \medskip
  As an English teacher, I need to prepare teaching materials about the word 'innovation'. \\
  & I need a detailed definition, sentence examples, pronunciation, and relevant pictures.  
  \medskip
\\ \hline

\multirow{12}{*}{\textbf{Action}} & 
  \medskip
  [ \\
  & \quad \{ \\
  & \quad \quad "name": "getWordDefinition", \\
  & \quad \quad "arguments": \{ "word": "innovation" \} \\
  & \quad \}, \\
  & \quad \{ \\
  & \quad \quad "name": "getWordExamples", \\
  & \quad \quad "arguments": \{ "word": "innovation" \} \\
  & \quad \}, \\
  & \quad \{ \\
  & \quad \quad "name": "getWordPronunciation", \\
  & \quad \quad "arguments": \{ "word": "innovation" \} \\
  & \quad \}, \\
  & \quad \{ \\
  & \quad \quad "name": "getWordImages", \\
  & \quad \quad "arguments": \{ "word": "innovation" \} \\
  & \quad \} \\
  & ]
  \medskip
\\ \hline
\multirow{40}{*}{\textbf{Observation}} &  
  \medskip
  [ \\
  & \quad \{ \\
  & \quad \quad "response": \{ \\
  & \quad \quad \quad "definitions": [ \\
  & \quad \quad \quad \quad \{ \\
  & \quad \quad \quad \quad \quad "definition": "The introduction of something new; a new idea, method, or device.", \\
  & \quad \quad \quad \quad \quad "partOfSpeech": "noun" \\
  & \quad \quad \quad \quad \}, \\
  & \quad \quad \quad \quad \{ \\
  & \quad \quad \quad \quad \quad "definition": "The act of innovating; the process of introducing new ideas, devices, or methods.", \\
  & \quad \quad \quad \quad \quad "partOfSpeech": "noun" \\
  & \quad \quad \quad \quad \} \\
  & \quad \quad \quad ], \\
  & \quad \quad \quad "word": "innovation" \\
  & \quad \quad \}, \\
  & \quad \quad "status\_code": 200 \\
  & \quad \}, \\
  & \quad \{ \\
  & \quad \quad "response": \{ \\
  & \quad \quad \quad "examples": [ \\
  & \quad \quad \quad \quad "The company's latest innovation is a new app that allows users to track their fitness progress.", \\
  & \quad \quad \quad \quad "Innovation in technology has dramatically changed the way we communicate.", \\
  & \quad \quad \quad \quad "The school encourages innovation in teaching methods to improve student engagement." \\
  & \quad \quad \quad ], \\
  & \quad \quad \quad "word": "innovation" \\
  & \quad \quad \}, \\
  & \quad \quad "status\_code": 200 \\
  & \quad \}, \\
  & \quad \{ \\
  & \quad \quad "response": \{ \\
  & \quad \quad \quad "images": [ \\
  & \quad \quad \quad \quad \{ \\
  & \quad \quad \quad \quad \quad "description": "A light bulb representing a new idea.", \\
  & \quad \quad \quad \quad \quad "type": "photo", \\
  & \quad \quad \quad \quad \quad "url": "https://example.com/images/innovation1.jpg" \\
  & \quad \quad \quad \quad \}, \\
  & \quad \quad \quad \quad \{ \\
  & \quad \quad \quad \quad \quad "description": "A group of people brainstorming new ideas.", \\
  & \quad \quad \quad \quad \quad "type": "photo", \\
  & \quad \quad \quad \quad \quad "url": "https://example.com/images/innovation2.jpg" \\
  & \quad \quad \quad \quad \}, \\
  & \quad \quad \quad \quad \{ \\
  & \quad \quad \quad \quad \quad "description": "A person using a new technology device.", \\
  & \quad \quad \quad \quad \quad "type": "photo", \\
  & \quad \quad \quad \quad \quad "url": "https://example.com/images/innovation3.jpg" \\
  & \quad \quad \quad \quad \} \\
  & \quad \quad \quad ], \\
  & \quad \quad \quad "word": "innovation" \\
  & \quad \quad \}, \\
  & \quad \quad "status\_code": 200 \\
  & \quad \}, \\
  & \quad \{ \\
  & \quad \quad "response": \{ \\
  & \quad \quad \quad "pronunciations": [ \\
  & \quad \quad \quad \quad \{ \\
  & \quad \quad \quad \quad \quad "audioFormat": "mp3", \\
  & \quad \quad \quad \quad \quad "url": "https://example.com/pronunciation/innovation.mp3" \\
  & \quad \quad \quad \quad \}, \\
  & \quad \quad \quad \quad \{ \\
  & \quad \quad \quad \quad \quad "audioFormat": "wav", \\
  & \quad \quad \quad \quad \quad "url": "https://example.com/pronunciation/innovation.wav" \\
  & \quad \quad \quad \quad \} \\
  & \quad \quad \quad ], \\
  & \quad \quad \quad "word": "innovation" \\
  & \quad \quad \}, \\
  & \quad \quad "status\_code": 200 \\
  & \quad \} \\
  & ]  
  \medskip
\\ \hline
\multirow{16}{*}{\textbf{Summary}} & 
  \medskip
  Here is the information for the word 'innovation':  
  \begin{itemize} 
    \item \textbf{Definition:}  
      - Noun: The introduction of something new (e.g., idea, method, device).  
      - Noun: The act of innovating or process of introducing new concepts/tools.  
    \item \textbf{Sentence Examples:}  
      ...
    \item \textbf{Pronunciation:}  
        ...
    \item \textbf{Images:}  
        ...
  \end{itemize}  
  \medskip
\\ \hline
\end{tabular}
\end{table*}

\begin{table*}[ht]
\centering
\small
\caption{Demo of policy $P_{multi}^{S + P}$. Data ID: 6ce42fc7-7ad6-4d27-8f08-e518d503a530
}
\label{demo-2-3}
\begin{tabular}{|c|p{12cm}|} 
\hline
\textbf{Role} & \textbf{Dialogue} \\
\hline
\multirow{8}{*}{\textbf{User}} & 
  \medskip
  I need some information about specific locations in Rwanda. Please search for a place called 'Kigali City Tower' and obtain its detailed information. Also, please provide a list of all districts in the Southern Province.
  \medskip
\\ \hline
\multirow{7}{*}{\textbf{Action}} & 
    \medskip
    [ \\
    & \quad \{ \\
      & \quad \quad "name": "getProvinces", \\
      & \quad \quad "arguments": \{ \} \\
      & \quad \}, \\
      & \quad \{ \\
      & \quad \quad "name": "searchLocation", \\
      & \quad \quad "arguments": \{ "query": "Kigali City Tower" \} \\
      & \quad \} \\ 
    & ]
    \medskip
\\ \hline
\multirow{9}{*}{\textbf{Observation}} & 
    \medskip
    [ \\
    & \quad \{ \\
      & \quad \quad "response": \{ \\
      & \quad \quad \quad "provinces": [\{ "id": 1, "name": "Kigali" \}, \{ "id": 2, "name": "Northern Province" \}, \{ "id": 3, "name": "Southern Province" \}, \{ "id": 4, "name": "Eastern Province" \}, \{ "id": 5, "name": "Western Province" \}] \\
      & \quad \quad \} \\
      & \quad \}, \\
      & \quad \{ \\
      & \quad \quad "response": \{ \\
      & \quad \quad \quad "locations": [\{ \\
      & \quad \quad \quad \quad "coordinates": \{ "latitude": - 1.9441, "longitude": 30.0619 \}, \\
      & \quad \quad \quad \quad "description": "Kigali City Tower is a famous landmark in Kigali, the capital of Rwanda.", \\
      & \quad \quad \quad \quad "id": 1, \\
      & \quad \quad \quad \quad "name": "Kigali City Tower", \\
      & \quad \quad \quad \quad "type": "building" \\
      & \quad \quad \quad ] \\
      & \quad \quad \} \\
      & \quad \} \\
    & ]
    \medskip
\\ \hline
\multirow{6}{*}{\textbf{Action}} & 
    \medskip
    [ \\
    & \quad \{ \\
      & \quad \quad "name": "getLocationDetails", \\
      & \quad \quad "arguments": \{ "locationId": 1 \} \\
      & \quad \}, \\
      & \quad \{ \\
      & \quad \quad "name": "getDistrictsByProvince", \\
      & \quad \quad "arguments": \{ "provinceId": 3 \} \\
      & \quad \} \\ 
    & ]
    \medskip
\\ \hline
\multirow{10}{*}{\textbf{Observation}} & 
    \medskip
    [ \\
    & \quad \{ \\
      & \quad \quad "response": \{ \\
      & \quad \quad \quad "districts": [\{ "id": 1, "name": "Huye" \}, \{ "id": 2, "name": "Nyanza" \}, \{ "id": 3, "name": "Gisagara" \}, \{ "id": 4, "name": "Nyaruguru" \}, \{ "id": 5, "name": "Nyamagabe" \}] \\
      & \quad \quad \}, \\
      & \quad \} \\
      & \quad \{ \\
      & \quad \quad "response": \{ \\
      & \quad \quad \quad "location": \{ \\
      & \quad \quad \quad \quad "address": "KN 2 St, Kigali, Rwanda", \\
      & \quad \quad \quad \quad "contact": \{ "email": "info@kigalitower.rw", "phone": "+250 788 123 456" \}, \\
      & \quad \quad \quad \quad "coordinates": \{ "latitude": - 1.9441, "longitude": 30.0619 \}, \\
      & \quad \quad \quad \quad "description": "Kigali City Tower is a famous landmark in Kigali, the capital of Rwanda. It is a multi - functional building that includes a shopping center, office space, and dining facilities.", \\
      & \quad \quad \quad \quad "id": 1, \\
      & \quad \quad \quad \quad "name": "Kigali City Tower", \\
      & \quad \quad \quad \quad "type": "building" \\
      & \quad \quad \} \\
      & \quad \} \\
    & ]
    \medskip
\\ \hline
\multirow{6}{*}{\textbf{Summary}} & 
    \medskip
  Here is the information you requested:  
  \begin{itemize} 
    \item Detailed information about Kigali City Tower:
      \begin{itemize}
        \item \textbf{Name}: Kigali City Tower
        \item \textbf{Description}: Kigali City Tower is a famous landmark in Kigali, ....
      \end{itemize}
  \end{itemize}  
  \medskip
\\ \hline
\end{tabular}
\end{table*}

\subsubsection{Three Type of Information Hidden Strategies}\label{demo-3}
Here, we show the demo of information hidden strategies, namely Omit, Reference, and long-context, 
in Tables \ref{demo-3-1}, \ref{demo-3-2}, \ref{demo-3-3}.

\begin{table*}[ht]
\centering
\caption{Demo of information hidden strategy: Omit. ID: 2c6ee79d-168d-45cf-a0d8-459243976d73}
\label{demo-3-1}
\begin{tabular}{|c|p{12cm}|} 
\hline
\textbf{Role} & \textbf{Dialogue} \\
\hline
\multirow{3}{*}{\textbf{User}} & 
  I want to travel to other places this weekend. \\
  & Please help me check the weather in Chicago \\
  & for the two days of the weekend.  
\\ \hline
\multirow{3}{*}{\textbf{Summary}} & 
  Here is the weather forecast for Chicago this weekend:  
  \begin{itemize} 
    \item On Saturday, July 13th, the weather will be light rain with a high of 25°C and a low of 18°C. The humidity will be 80\%.  
    \item On Sunday, July 14th, the weather will be moderate rain with a high of 23°C and a low of 17°C. The humidity will be 85\%.  
  \end{itemize}  
\\ \hline

\multirow{1}{*}{\textbf{User}} & 
  It doesn't look so good. Please help me check Las Vegas again?
 
\\ \hline

\end{tabular}
\end{table*}

\begin{table*}[ht]
\centering
\caption{Demo of information hidden strategy: Reference. ID: 00d623ee-700d-4f6a-a26f-00df57b2ccdd}
\label{demo-3-2}
\begin{tabular}{|c|p{12cm}|} 
\hline
\textbf{Role} & \textbf{Dialogue} \\
\hline
\multirow{3}{*}{\textbf{User}} & 
  I want to sell some RuneScape game items, specifically the dragon helm with item ID 2513. I have 20 of them. Can you help me with this?  
\\ \hline
\multirow{4}{*}{\textbf{Summary}} & 
  The item you requested has been successfully sold. The specific information is as follows:  
  \begin{itemize} 
    \item \textbf{Item ID}: 2513
    \item \textbf{Quantity}: 20
    \item \textbf{Status}: Success
    \item \textbf{Total Revenue}: 100,000 coins
  \end{itemize}  
\\ \hline
\multirow{1}{*}{\textbf{User}} & 
  Sell 10 more.
\\ \hline

\end{tabular}
\end{table*}

\begin{table*}[ht]
\centering
\caption{Demo of information hidden strategy: Long Context. ID: bdf7dfee-8bf6-417a-9ad1-07309e4e2460}
\label{demo-3-3}
\begin{tabular}{|c|p{12cm}|} 
\hline
\textbf{Role} & \textbf{Dialogue} \\
\hline
\multirow{2}{*}{\textbf{User}} & 
  There is a station in northern Chicago. I need to know about its wheelchair access situation.  
\\ \hline
\multirow{1}{*}{\textbf{Summary}} & 
  Could you please provide the specific name or identifier of the station you want to know about?  
\\ \hline
\multirow{2}{*}{\textbf{User}} & 
  I want to know about the wheelchair access at Howard Station in northern Chicago.  
\\ \hline
\multirow{2}{*}{\textbf{Summary}} & 
  Howard Station in northern Chicago is wheelchair accessible. \\
  & The operating hours of the station are from 5 am to 1 am.  
\\ \hline
\multirow{5}{*}{\textbf{User}} & 
  Are there any other stations with wheelchair access? \\
  & \quad In addition to Howard Station, there are many stations with wheelchair access in the northern part of Chicago. Here are some listed for you: \\
  & \quad \begin{itemize} 
    \item \textbf{Purple Line}: Linden, Davis.  
    \item \textbf{Red Line}: Loyola, Granville.  
    \item \textbf{Yellow Line}: Oakton - Skokie, Dempster - Skokie.  
  \end{itemize} \\
  & \quad In addition, there are many stations with wheelchair access on other lines and in other areas of Chicago, such as O’Hare, Rosemont, and Cumberland on the Blue Line; Kimball, Kedzie, Francisco on the Brown Line, etc. \\
  & What are the train platforms at the terminal station to reach there? At Chicago General Station, the current train platform information is as follows: \\
  & \quad \begin{itemize} 
    \item \textbf{Platform 1}: Metra UP - North Line, direction is arrivals.  
    \item \textbf{Platform 2}: Amtrak Cardinal, direction is departures.  
    \item \textbf{Platform 3}: Metra BNSF Line, direction is arrivals.  
  \end{itemize}  
\\ \hline
\multirow{1}{*}{\textbf{User}} & 
  Check the train schedule there again.  
\\ \hline
\end{tabular}
\end{table*}

\subsubsection{Policy Switching}\label{demo-4}
We show a demo of policy switching multi-task in Tables \ref{demo-4-1},  \ref{demo-4-2},  \ref{demo-4-3},  \ref{demo-4-4}. They are sequential 4 tasks, and have different policies.

\begin{table*}[ht]
\centering
\caption{Demo for 4 time policy switching, Task 1. The corresponding policies are: $P_{\text{single}}, P_{\text{multi}}, P_{\text{clarify}}, P_{\text{chat}}$. }
\label{demo-4-1}
\begin{tabular}{|c|p{12cm}|} 
\hline
\textbf{Role} & \textbf{Dialogue} \\
\hline

\multirow{3}{*}{\textbf{User}} & 
  Request to overlay polygon data on map with ID "GlobalMap". Coordinates: (48.8566, 2.3522) and (41.9028, 12.4964), transparency set to 50\%.\\ \hline
\multirow{7}{*}{\textbf{Action}} & 
    [ \\
    & \quad \{ \\
      & \quad \quad "name": "overlayDataOnMap", \\
      & \quad \quad "arguments": \{ \\
      & \quad \quad \quad "Map ID": "GlobalMap", \\
      & \quad \quad \quad "Data type": "polygon", \\
      & \quad \quad \quad "Coordinates": [\{ \\
      & \quad \quad \quad \quad "latitude": 48.8566, \\
      & \quad \quad \quad \quad "longitude": 2.3522 \\
      & \quad \quad \quad \}, \{ \\
      & \quad \quad \quad \quad "latitude": 41.9028, \\
      & \quad \quad \quad \quad "longitude": 12.4964 \\
      & \quad \quad \quad \}], \\
      & \quad \quad \quad "Properties": \{ "opacity": 0.5 \} \\
      & \quad \quad \} \\
      & \quad \} \\ 
    & ]
\\ \hline
\multirow{10}{*}{\textbf{Observation}} &  
    [ \\
    & \quad \{ \\
      & \quad \quad "status": 200, \\
      & \quad \quad "message": "The polygon data has been successfully overlaid on the map with map ID 'GlobalMap'.", \\
      & \quad \quad "details": \{ \\
      & \quad \quad \quad "Map ID": "GlobalMap", \\
      & \quad \quad \quad "Data type": "polygon", \\
      & \quad \quad \quad "Coordinates": [\{ \\
      & \quad \quad \quad \quad "latitude": 48.8566, \\
      & \quad \quad \quad \quad "longitude": 2.3522 \\
      & \quad \quad \quad \}, \{ \\
      & \quad \quad \quad \quad "latitude": 41.9028, \\
      & \quad \quad \quad \quad "longitude": 12.4964 \\
      & \quad \quad \quad \}], \\
      & \quad \quad \quad "Properties": \{ "opacity": 0.5 \} \\
      & \quad \quad \} \\
      & \quad \} \\ 
    & ]
\\ \hline
\multirow{1}{*}{\textbf{Summary}} & 
  The polygon data has been successfully overlaid on the map with ID "GlobalMap", with transparency set to 50\%.  
\\ \hline
\end{tabular}
\end{table*}

\begin{table*}[ht]
\centering
\caption{Demo for 4 time policy switching, Task 2. The corresponding policies are: $P_{\text{single}}, P_{\text{multi}}, P_{\text{clarify}}, P_{\text{chat}}$. }
\label{demo-4-2}
\begin{tabular}{|c|p{12cm}|} 
\hline
\textbf{Role} & \textbf{Dialogue} \\
\hline
\textbf{User} & Please mark these two coordinate points with titles Statue of Liberty - Northwest and Statue of Liberty - Southeast respectively for me. \\ \hline
\multirow{7}{*}{\textbf{Action}} & 
    [ \\
    & \quad \{ \\
      & \quad \quad "name": "addCustomMarker", \\
      & \quad \quad "arguments": \{ \\
      & \quad \quad \quad "mapId": "GlobalMap", \\
      & \quad \quad \quad "position": \{ \\
      & \quad \quad \quad \quad "latitude": 48.8566, \\
      & \quad \quad \quad \quad "longitude": 2.3522 \\
      & \quad \quad \quad \}, \\
      & \quad \quad \quad "title": "Statue of Liberty - Northwest" \\
      & \quad \quad \} \\
      & \quad \}, \\
      & \quad \{ \\
      & \quad \quad "name": "addCustomMarker", \\
      & \quad \quad "arguments": \{ \\
      & \quad \quad \quad "mapId": "GlobalMap", \\
      & \quad \quad \quad "position": \{ \\
      & \quad \quad \quad \quad "latitude": 41.9028, \\
      & \quad \quad \quad \quad "longitude": 12.4964 \\
      & \quad \quad \quad \}, \\
      & \quad \quad \quad "title": "Statue of Liberty - Southeast" \\
      & \quad \quad \} \\
      & \quad \} \\ 
    & ]
\\ \hline
\multirow{7}{*}{\textbf{Observation}} &  
    [ \\
    & \quad \{ \\
      & \quad \quad "response": \{ \\
      & \quad \quad \quad "message": "The marker has been successfully added to the location (48.8566, 2.3522) on the map with ID 'GlobalMap' and the title 'Statue of Liberty - Northwest'." \\
      & \quad \quad \} \\
      & \quad \}, \\
      & \quad \{ \\
      & \quad \quad "response": \{ \\
      & \quad \quad \quad "message": "The marker has been successfully added to the location (41.9028, 12.4964) on the map with ID 'GlobalMap' and the title 'Statue of Liberty - Southeast'." \\
      & \quad \quad \} \\
      & \quad \} \\ 
    & ]
\\ \hline
\textbf{Summary} & 
  Operation Results: Two markers have been successfully added to the map with ID "GlobalMap": 
  \begin{itemize} 
    \item \textbf{Statue of Liberty - Northwest}: Located at latitude 48.8566 and longitude 2.3522.
    \item \textbf{Statue of Liberty - Southeast}: Located at latitude 41.9028 and longitude 12.4964.
  \end{itemize}
  The markers have been successfully added and you can view these locations on the map.\\ \hline
\end{tabular}
\end{table*}

\begin{table*}[ht]
\centering
\caption{Demo for 4 time policy switching, Task 3. The corresponding policies are: $P_{\text{single}}, P_{\text{multi}}, P_{\text{clarify}}, P_{\text{chat}}$. }
\label{demo-4-3}
\begin{tabular}{|c|p{12cm}|} 
\hline
\textbf{Role} & \textbf{Dialogue} \\
\hline
\textbf{User} & Mark one more for me. \\ \hline
\textbf{Summary} & Please provide the geographical coordinates (latitude and longitude) of the marker and the title of the marker. \\ \hline
\textbf{User} & 122.4194, San Francisco. \\ \hline
\textbf{Summary} & Please provide the longitude of the marker. \\ \hline
\textbf{User} & 37.7749. \\ \hline
\multirow{7}{*}{\textbf{Action}} & 
    [ \\
    & \quad \{ \\
      & \quad \quad "name": "addCustomMarker", \\
      & \quad \quad "arguments": \{ \\
      & \quad \quad \quad "mapId": "GlobalMap", \\
      & \quad \quad \quad "position": \{ \\
      & \quad \quad \quad \quad "latitude": 122.4194, \\
      & \quad \quad \quad \quad "longitude": 37.7749 \\
      & \quad \quad \quad \}, \\
      & \quad \quad \quad "title": "San Francisco" \\
      & \quad \quad \} \\
      & \quad \} \\ 
    & ]
\\ \hline
\multirow{7}{*}{\textbf{Observation}} &  
    [ \\
    & \quad \{ \\
      & \quad \quad "response": \{ \\
      & \quad \quad \quad "city": "San Francisco", \\
      & \quad \quad \quad "country": "United States", \\
      & \quad \quad \quad "latitude": 37.7749, \\
      & \quad \quad \quad "longitude": 122.4194, \\
      & \quad \quad \quad "region": "California" \\
      & \quad \}, \\
      & \quad \quad "status\_code": 200 \\
      & \quad \} \\ 
    & ]
\\ \hline
\textbf{Summary} & 
  Operation Results  
  \begin{itemize} 
    \item A marker has been successfully added on the map with ID "GlobalMap":
      \begin{itemize}
        \item \textbf{San Francisco}: Located at latitude 37.7749 and longitude 122.4194.
      \end{itemize}
  \end{itemize}
  The marker has been successfully added. You can view these locations on the map.\\ \hline
\end{tabular}
\end{table*}

\begin{table*}[ht]
\centering
\caption{Demo for 4 time policy switching, Task 4. The corresponding policies are: $P_{\text{single}}, P_{\text{multi}}, P_{\text{clarify}}, P_{\text{chat}}$. }
\label{demo-4-4}
\begin{tabular}{|c|p{12cm}|} 
\hline
\textbf{Role} & \textbf{Dialogue} \\
\hline
\textbf{User} & Help me add this map to my mapping software \\ \hline
\textbf{Summary} & I understand you want to add the map configuration we've been working on (with ID "GlobalMap") to your mapping software. However, the available tools do not support exporting the map data or linking it directly to external mapping software. My capabilities are limited to actions like embedding maps, adding markers, and overlaying data using the provided functions. \\ \hline
\end{tabular}
\end{table*}

\section{Error Cases}\label{chap:err}

In this chapter, we present several typical error examples discussed in the main text.

\begin{table*}[ht]
\centering
\caption{The different performances of multiple models on a multi-tool task.}
\label{demo-19}
\begin{tabular}{|c|p{12cm}|} 
\hline
\textbf{Role} & \textbf{Dialogue} \\
\hline
\multirow{1}{*}{\textbf{User}} & 
  As a pilot, I need to obtain the current weather conditions at San Francisco International Airport, and please include pilot report data.  
\\ \hline
\multirow{1}{*}{\textbf{Summary}} & 
  Current Weather Conditions at San Francisco International Airport (SFO)
  METAR Report  
  - Observation Time: 2024-10-09 23:53 UTC  
  - Temperature: 18.3°C  
  - Dew Point: 11.7°C  
  - Wind: 280°, 15 knots  
  - Visibility: 10 statute miles  
  - Cloud Cover: Few clouds, cloud base at 2000 feet  
  - Altimeter: 30.05 inches of mercury 

  Please note that this information may be helpful in flight planning and safety decisions.  
\\ \hline
\multirow{1}{*}{\textbf{User}} & 
  Weather alerts, delay information?  
\\ \hline
\multirow{3}{*}{ \textbf{GPT-4.1} \textcolor{teal}{\Checkmark} } & 
    [ \\
    & \quad \{ \\
      & \quad \quad "name": "getAirportDelays", \\
      & \quad \quad "arguments": \{ \\
      & \quad \quad \quad "airportCode": "SFO" \\
      & \quad \quad \} \\
      & \quad \}, \\
      & \quad \{ \\
      & \quad \quad "name": "getWeatherAlerts", \\
      & \quad \quad "arguments": \{ \\
      & \quad \quad \quad "region": "SFO" \\
      & \quad \quad \} \\
      & \quad \} \\ 
    & ]
\\ \hline

\multirow{1}{*}{\textbf{Gemini-2.0-thinking}\textcolor{red}{\XSolidBrush}} &  
  I can help you with that. To get the weather alerts, I need to know the specific region you are interested in. For the delay information, which airport are you asking about? If it's San Francisco International Airport (SFO), I can look that up, but please confirm or provide the airport code.  
\\ \hline
\multirow{3}{*}{ \textbf{Gemini-2.5-pro}\textcolor{red}{\XSolidBrush}} & 
    [ \\
    & \quad \{ \\
      & \quad \quad "name": "getAirportDelays", \\
      & \quad \quad "arguments": \{ \\
      & \quad \quad \quad "airportCode": "SFO" \\
      & \quad \quad \} \\
      & \quad \} \\ 
    & ]
\\ \hline
\multirow{1}{*}{\textbf{o1}\textcolor{red}{\XSolidBrush}} &  
  Could you please specify the region code for which you need weather alerts? For example, if you need SIGMET or AIRMET information for the Oakland Air Route Traffic Control Center area, you might use “ZOA” as the region code. Also please let me know if you have a preference for specific delay types (e.g., weather delays) or if you’d like to see all delay information.  
\\ \hline
\end{tabular}
\end{table*}

\begin{table*}[ht]
\centering
\caption{Falth to asking users to repeat the previous dialogue.}
\label{demo-20}
\begin{tabular}{|c|p{12cm}|} 
\hline
\textbf{Role} & \textbf{Dialogue} \\
\hline
\multirow{1}{*}{\textbf{User}} & 
  As a pilot, I need to obtain the current weather conditions at San Francisco International Airport, and please include pilot report data.  
\\ \hline
\multirow{1}{*}{\textbf{Summary}} & 
  Current Weather Conditions at San Francisco International Airport (SFO)
  METAR Report  
  - Observation Time: 2024-10-09 23:53 UTC  
  - Temperature: 18.3°C  
  - Dew Point: 11.7°C  
  - Wind: 280°, 15 knots  
  - Visibility: 10 statute miles  
  - Cloud Cover: Few clouds, cloud base at 2000 feet  
  - Altimeter: 30.05 inches of mercury

  Please note that this information may be helpful in flight planning and safety decisions.  
\\ \hline
\multirow{1}{*}{\textbf{User}} & 
  Weather alerts, delay information?  
\\ \hline
\multirow{3}{*}{ \textbf{GPT-4.1} \textcolor{teal}{\Checkmark} } & 
    [ \\
    & \quad \{ \\
      & \quad \quad "name": "getAirportDelays", \\
      & \quad \quad "arguments": \{ \\
      & \quad \quad \quad "airportCode": "SFO" \\
      & \quad \quad \} \\
      & \quad \}, \\
      & \quad \{ \\
      & \quad \quad "name": "getWeatherAlerts", \\
      & \quad \quad "arguments": \{ \\
      & \quad \quad \quad "region": "SFO" \\
      & \quad \quad \} \\
      & \quad \} \\ 
    & ]
\\ \hline

\multirow{1}{*}{\textbf{Gemini-2.0-thinking}\textcolor{red}{\XSolidBrush}} &  
  I can help you with that. To get the weather alerts, I need to know the specific region you are interested in. For the delay information, which airport are you asking about? If it's San Francisco International Airport (SFO), I can look that up, but please confirm or provide the airport code.  
\\ \hline

\end{tabular}
\end{table*}

\end{document}